\documentclass{article}

\usepackage{neurips_2026}

\usepackage[utf8]{inputenc} 
\usepackage[T1]{fontenc}    
\usepackage[pagebackref,breaklinks,colorlinks,citecolor=green]{hyperref}
\usepackage{url}            
\usepackage{booktabs}       
\usepackage{amsfonts}       
\usepackage{nicefrac}       
\usepackage{microtype}      
\usepackage{xcolor}         
\usepackage{diagbox}
\usepackage{booktabs}
\usepackage{tabularx}
\usepackage{amssymb}
\usepackage{multirow}
\usepackage{makecell}
\usepackage{threeparttable}
\usepackage{pifont} 
\usepackage[table]{xcolor}
\usepackage{xspace}  
\usepackage{graphicx}
\usepackage{wrapfig}
\usepackage{changepage}
\usepackage{setspace}
\usepackage{amsmath}

\newcommand{\redcell}[1]{{\setlength{\fboxsep}{1pt}\colorbox{red!15}{#1}}}
\newcommand{\yellowcell}[1]{{\setlength{\fboxsep}{1pt}\colorbox{yellow!25}{#1}}}
\newcommand{\bluecell}[1]{{\setlength{\fboxsep}{1pt}\colorbox{blue!15}{#1}}}
\newcommand{\graycell}[1]{{\setlength{\fboxsep}{1pt}\colorbox{gray!25}{#1}}}
\title{COVD: Continual Open-Vocabulary Object Detection \\ with Novel Concept Injection}

\author{
	Yupeng Zhang\textsuperscript{\rm 1}\hspace{1em}
	Ruize Han\textsuperscript{\rm 2}\hspace{1em}
	Yuzhong Feng\textsuperscript{\rm 1}\hspace{1em}
    Zixin Ren\textsuperscript{\rm 1}\hspace{1em}
    Yuntong Tian\textsuperscript{\rm 1}\hspace{1em}
	Liang Wan\textsuperscript{\rm 1}\hspace{1em}\\  
	\textsuperscript{\rm 1}Tianjin University.~~~~
	\textsuperscript{\rm 2}Shenzhen University of Advanced Technology.\\
{\tt\small \{zhangyupeng, yzfeng, renzixin, tianyuntong, lwan\}@tju.edu.cn, hanruize@suat-sz.edu.cn}
}

\begin{document}

\maketitle

\begin{abstract}
Open-vocabulary object detection (OVD) has made significant progress, enabling detectors to generalize from seen to unseen categories. However, real-world category spaces continually evolve, and existing OVD models still struggle with newly emerging concepts, while repeated full retraining is prohibitively expensive.
To this end, we introduce a new task setting, termed Continual OVD with Novel Concept Injection (COVD), 
where models sequentially learn incoming novel concept groups while preserving prior concepts and original open-vocabulary knowledge, along with a new benchmark, Novel-114.
Our key observation is that pretrained visual encoders often already perceive and represent many novel concepts, and the main bottleneck lies in the lack of stable semantic alignment between visual representations and textual concepts. Based on this, we propose NoIn-Det, an efficient continual injection framework without additional parameters. NoIn-Det freezes the visual encoder, preserves the text representation space using only texts of common concepts and previously injected concepts, and injects novel concepts by updating only a small subset of text-branch parameters beneficial to novel concept learning.
Extensive experiments show that NoIn-Det effectively learns novel concepts, preserves old knowledge, and consistently outperforms existing continual learning methods for VLMs without introducing additional parameters. \textit{Novel-114 and the code will be released.}
\end{abstract}

\section{Introduction}
\label{sec:intro}

\begin{wrapfigure}{r}{0.51\textwidth}\vspace{-12pt}
	\centering
	\includegraphics[width=1.0\linewidth]{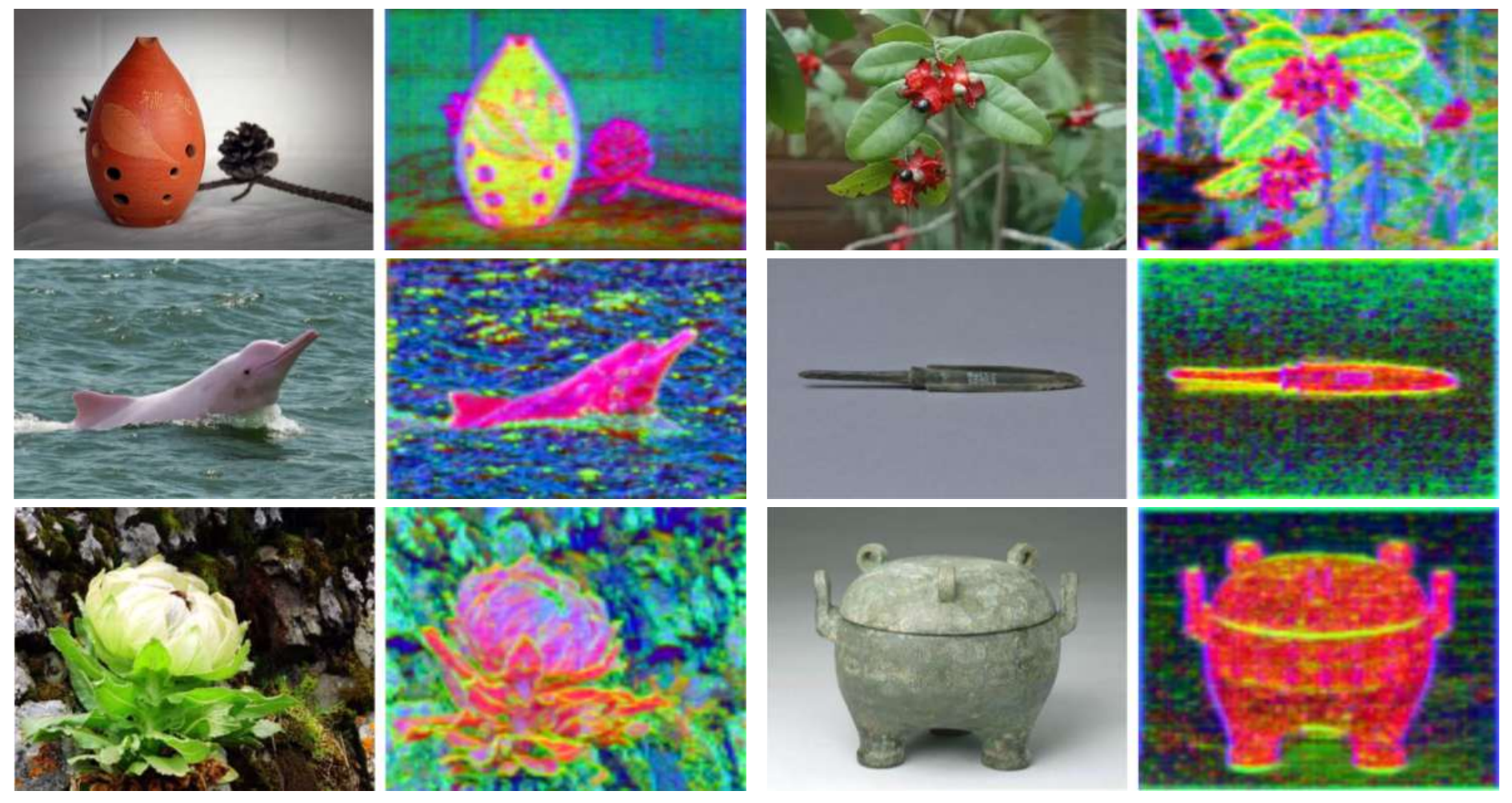}
	\vspace{-22pt}
	\caption{Visualization of features from the pretrained visual backbone. Backbone outputs are projected to RGB space using PCA. Although the objects are novel concepts unseen during pretraining, the backbone highlights object regions and produces coherent responses within semantic areas.} 
	\label{fig:PCA}
	\vspace{-15pt}
\end{wrapfigure}

In recent years, open-vocabulary object detection (OVD) has extended object detection beyond fixed category sets by leveraging the textual semantic space learned from large-scale vision-language pre-training and the ability to recombine known semantic concepts.
However, real-world category spaces evolve continuously. New concepts emerge across scenarios, domains, and tasks, including long-tail categories with scarce data, specialized terms, and domain-specific objects. While existing OVD methods can generalize to conventional unseen categories inferred from known semantics, they still struggle to establish reliable vision-language alignment for \textbf{genuinely novel concepts}, leading to degraded detection performance.
When novel concepts emerge continually, full retraining can absorb new knowledge and preserve existing capabilities, but requires large-scale historical image-text data and prohibitive training costs, which further grow as concepts accumulate. Therefore, enabling open-vocabulary detectors to continually learn novel concepts without repeated full retraining, while preserving both learned concepts and original open-vocabulary knowledge, becomes an important and practical problem.

Based on the above observations, we introduce a new task, \textbf{Continual Open-Vocabulary Object Detection with Novel Concept Injection (COVD)}, which aims to enable open-vocabulary detectors to continually absorb novel concepts after deployment in a parameter-efficient and progressive manner, without repeated full retraining. In this setting, the model sequentially learns incoming groups of novel concepts and is evaluated at each stage on current concept acquisition, previously injected concept retention, and preservation of original open-vocabulary knowledge. Unlike ZiRa~\cite{deng2024zero}, which focuses on incremental adaptation to different downstream tasks while preserving zero-shot generalization, COVD emphasizes concept-level continual learning, aiming to inject novel concepts while retaining both previously injected concepts and the original open-vocabulary knowledge.
Moreover, existing continual learning benchmarks for VLM-based object detection are insufficient for evaluating genuine novel concept injection. For example, ODinW-13~\cite{li2022elevater}, used in ZiRa~\cite{deng2024zero}, mainly introduces common categories already likely covered by large-scale vision-language pre-training. As a result, gains on such benchmarks may reflect better use of existing semantic priors rather than true acquisition of novel concepts. To address this, we construct \textbf{Novel-114}, a benchmark with 114 more challenging novel concepts organized into 7 scenario-based continual injection groups, enabling systematic evaluation of both novel concept acquisition and retention of previously injected concepts.

Although continual learning for VLMs has been studied, existing methods still have clear limitations.
1) Prompt-based methods~\cite{tang2024mind} often store novel knowledge in external prompts rather than truly teaching the model novel concepts, and require these prompts to be preserved and managed as concepts accumulate.
2) Adapter-based methods~\cite{cheng2026incremental,luo2025lada} and mixture-of-experts methods~\cite{yu2024boosting} improve adaptation, but introduce extra parameters and continuously enlarge the model.
3) Parameter-efficient fine-tuning~\cite{luo2026keeplora} and reparameterization~\cite{deng2024zero} may also disrupt the original semantic space during novel concept injection, leading to unstable old knowledge representations.
\textbf{Overall, existing methods rely on external modules or parameter adaptation for short-term adjustment, rather than truly and stably internalizing novel concepts into the model, making it difficult to balance novel concept learning and old knowledge retention.} \textbf{\textit{See Appendix~\ref{sec:Related} for additional related work discussion.}}

\begin{wrapfigure}{r}{0.5\textwidth}\vspace{-18pt}
	\centering
	\includegraphics[width=1.0\linewidth]{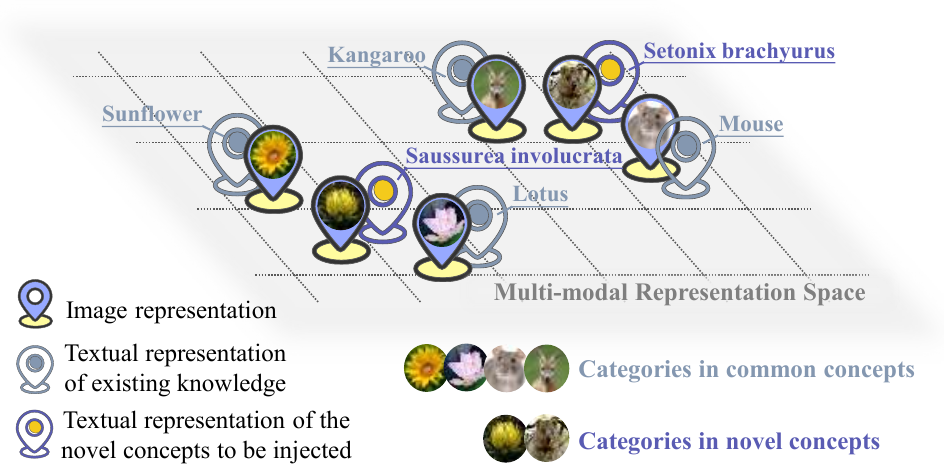}
	\vspace{-20pt}
	\caption{Illustration of the multi-modal representation space. After large-scale pre-training, the visual encoder of a VLM can represent visual cues such as color, material, texture, shape, and local structure, and map novel concepts to reasonable positions in the multi-modal space based on their correlations. For example, Setonix brachyurus shares similar fur color, texture, and overall appearance with a kangaroo, while its frontal local shape is closer to that of a mouse; therefore, its visual representation is closer to those of kangaroos and mice.} 
	\label{fig:over}
	\vspace{-10pt}
\end{wrapfigure}

More importantly, as shown in Fig.~\ref{fig:PCA}, pretrained visual encoders exhibit strong perception and representation ability for many novel concepts. Since novel objects are often composed of familiar visual primitives, such as colors, materials, textures, shapes, and local structures, the encoder can map them into a reasonable visual representation space even if they are not explicitly seen during training, as illustrated in Fig.~\ref{fig:over}. Therefore, the key limitation is not the lack of visual perception of novel objects, but the absence of stable and accurate semantic alignment between their visual representations and corresponding textual concepts. \textbf{In other words, the model can `see' novel objects but has not learned their `names' in human semantics.} 
This observation further suggests that the key to continual novel concept injection is not to reshape visual representations~\cite{cheng2026incremental} or jointly update both vision and language branches~\cite{deng2024zero,yu2024boosting,liu2025c,luo2026keeplora}, as in existing VLMs continual learning methods. Instead, it should establish semantic bindings between novel concepts and their corresponding visual representations from the textual side, while keeping the visual space as stable as possible. This preserves the general perception and representation ability acquired through large-scale pre-training, avoids unnecessary perturbation to complex visual patterns, and improves the stability and interpretability of novel concept learning.

Motivated by the above, we propose NoIn-Det, an efficient continual novel concept injection framework for OVD. NoIn-Det introduces no extra parameters and freezes the visual encoder, injecting novel concepts by only fine-tuning key parameters in the text branch, while using textual constraints from common concepts and previously injected concepts to maintain a stable representation space.
Specifically, NoIn-Det first learns current novel concepts through contrastive learning on their corresponding region-text pairs, providing basic absorption and alignment of new knowledge. On this basis, to preserve the stability of the textual representation space, we design \textbf{Representation Space Stability Distillation (RSSD)}, which uses the original textual branch as a teacher and performs both absolute semantic distillation and relative relational distillation on randomly sampled common concepts and stored previously injected concepts. Since text is easy to obtain and store, this strategy provides continual old-knowledge regularization at low cost.
Meanwhile, to mitigate the conflict between novel concept learning and old knowledge retention, we introduce \textbf{Knowledge-aware Parameter Decoupling (KPD)}, which suppresses excessive drift of parameters sensitive to old knowledge and concentrates updates on parameters more beneficial to novel concept learning.
Unlike existing continual learning methods for VLMs that rely on historical image-text data, separate modules, external prompts, or parameter adapters to carry new knowledge, NoIn-Det emphasizes stably injecting novel concepts into the existing semantic space without expanding the model architecture.

Our contributions are summarized as follows:
\begin{itemize}
\item We introduce COVD, a continual novel concept injection task for OVD, and construct Novel-114 to evaluate novel concept acquisition and retention of previously injected concepts.
\item We propose NoIn-Det, an efficient continual injection framework without additional parameters. By freezing the visual encoder, updating only key textual parameters, and using RSSD and KPD to preserve textual representation stability and mitigate novel-old knowledge conflicts, NoIn-Det enables injection of novel concepts into the existing semantic space.
\item Experiments on Novel-114 and MS COCO show that NoIn-Det effectively improves novel concept learning and outperforms existing continual learning methods for VLMs in retaining both previously injected concepts and original open-vocabulary knowledge.
\end{itemize}

\section{Novel-114: A Benchmark for COVD}
\label{sec:Benchmark}

\textbf{Motivation.}
Existing continual learning benchmarks for detection VLMs are insufficient for evaluating continual novel concept injection in OVD. For example, ODinW-13~\cite{li2022elevater}, used in ZiRa~\cite{deng2024zero}, mainly introduces common concepts covered by large-scale vision-language pre-training, and contains clear domain shifts such as infrared, grayscale, and underwater imagery. This matches ZiRa's focus on gradual adaptation to specialized scenarios, tasks, or domains. Therefore, gains on such benchmarks may reflect better use of existing semantic priors or domain adaptation, rather than genuine novel concept learning.
Similarly, although LVIS~\cite{gupta2019lvis} contains rare categories, their number is limited, and many are supplementary annotations built on COCO, often involving small objects, severe occlusion, and sparse instances. Thus, their difficulty may arise more from perceptual challenges than from concept learning.
To this end, we construct Novel-114, a new benchmark for COVD. Unlike existing benchmarks, Novel-114 focuses on concepts insufficiently covered by vision-language pre-training corpora and common detection benchmarks, enabling stricter evaluation of whether models acquire new conceptual knowledge rather than transfer from existing semantic priors.

\textbf{Design principles.}
In constructing Novel-114, we follow five principles:
\textbf{\ding{172} \textit{Novelty}:} Concepts are not well covered by common detection benchmarks or large-scale pre-training corpora.
\textbf{\ding{173} \textit{Learnability}:} Extreme small objects, severe occlusion, and harsh imaging conditions are minimized to focus evaluation on concept learning.
\textbf{\ding{174} \textit{Continuity}:} Concepts support multi-stage incremental injection to evaluate both knowledge acquisition and retention.
\textbf{\ding{175} \textit{Diversity}:} Groups show large scenario gaps and varied concept numbers, enabling evaluation under diverse learning scenarios.
\textbf{\ding{176} \textit{Annotation reliability}:} Instances are carefully annotated and manually verified to ensure clear definitions, accurate boxes, and consistent train/test labels.
Based on these principles, we collect images from multiple domains and use CLIP~\cite{radford2021learning} and LLMDet~\cite{fu2025llmdet} to screen candidate categories, filtering out concepts already well covered by existing VLMs and OVD models. We construct Novel-114 with 114 categories, organized into 7 continual learning groups by scenario and semantic relevance, each corresponding to one learning stage. The group names indicate the number of concepts, e.g., Animals-25 and Clothing-6 contain 25 animal and 6 clothing concepts, respectively.
We adopt a fixed heterogeneous injection order: Animals-25, Clothing-6, Cold Weapons-4, Food-28, Instruments-8, Plants-25, and Vessels-18. This order is not based on category number, visual difficulty, or semantic similarity, but alternates across natural objects, man-made objects, specialized concepts, and daily-life concepts, with group sizes ranging from 4 to 28. This design simulates realistic scenarios where novel concepts arrive from diverse domains and batch sizes, enabling evaluation under non-curriculum, cross-scenario, and non-uniform continual learning settings.

\begin{wraptable}{r}{0.5\textwidth} \vspace{-19pt}
\centering
\scriptsize
\setlength{\tabcolsep}{4.0pt}
\caption{Statistics of each task in the Novel-114.}\vspace{1pt}
\label{tab:novel114}
\begin{tabular}{lcccc}
\hline
\textbf{Task} & \textbf{\# Concepts} & \textbf{Subset} & \textbf{\# Images} & \textbf{\# Objects} \\
\hline
\multirow{2}{*}{Animals-25} & \multirow{2}{*}{25} & train & 1049 & 1334 \\
 &  & test & 1057 & 1360 \\\hline
\multirow{2}{*}{Clothing-6} & \multirow{2}{*}{6} & train & 184 & 211 \\
 &  & test & 187 & 231 \\\hline
\multirow{2}{*}{Cold Weapons-4} & \multirow{2}{*}{4} & train & 150 & 272 \\
 &  & test & 151 & 247 \\\hline
\multirow{2}{*}{Food-28} & \multirow{2}{*}{28} & train & 1407 & 3093 \\
 &  & test & 1420 & 3188 \\\hline
\multirow{2}{*}{Instruments-8} & \multirow{2}{*}{8} & train & 134 & 170 \\
 &  & test & 139 & 250 \\\hline
\multirow{2}{*}{Plants25} & \multirow{2}{*}{25} & train & 873 & 3456 \\
 &  & test & 882 & 3370 \\\hline
\multirow{2}{*}{Vessels-18} & \multirow{2}{*}{18} & train & 627 & 757 \\
 &  & test & 631 & 759 \\
\hline
\multirow{2}{*}{\textbf{All Categories}} & \multirow{2}{*}{\textbf{114}} & \textbf{train} & \textbf{4424} & \textbf{9293} \\
 &  & \textbf{test} & \textbf{4467} & \textbf{9405} \\
\hline
\end{tabular}\vspace{-15pt}
\end{wraptable}
\textbf{Statistics.}
As shown in Table~\ref{tab:novel114}, Novel-114 contains 114 novel concepts organized into 7 continual learning tasks. Each task is split into training and test sets, yielding 4,424 training images with 9,293 annotations and 4,467 test images with 9,405 annotations in total. The tasks vary substantially in concept count, image scale, and object instances, forming a continual benchmark with a clearly non-uniform task distribution. 
Unlike conventional static detection benchmarks, Novel-114 is naturally organized for continual learning: the model learns novel concepts task by task, and is evaluated at each stage on both current concept acquisition and retention of previously injected concepts. 
Therefore, Novel-114 provides a more targeted and challenging benchmark for continual novel concept injection, enabling comprehensive evaluation of the balance between novel concept learning and old knowledge retention.
\textbf{\textit{More details of Novel-114 are in Appendix~\ref{sec:More_details_novel114}.}}

\section{The Proposed Baseline Method} 
\label{sec:Method}

\subsection{Problem Formulation}
We first formalize the proposed \textbf{COVD} problem.
At initialization, we are given an open-vocabulary detector {\footnotesize $f_{\theta_0}$}.
During continual learning, the model sequentially learns a series of novel concept sets, denoted by {\footnotesize $\{\mathcal{C}^{t}\}_{t=1}^{T}$}, where {\footnotesize $T$} is the total number of continual learning tasks, {\footnotesize $\mathcal{C}^{t}$} denotes the set of novel concepts introduced at task {\footnotesize $t$}, and {\footnotesize $\mathcal{C}^{i}\cap\mathcal{C}^{j}=\varnothing$ for any {\footnotesize $i\neq j$}}.
For each task {\footnotesize $t\in\{1,\dots,T\}$}, the corresponding training set is defined as
{\footnotesize $\mathcal{D}^{t}=\{(\mathbf{I}_i^{t},L_i^{t})\}_{i=1}^{N_t}$},
where {\footnotesize $N_t$} is the number of training images in task {\footnotesize $t$}, {\footnotesize $\mathbf{I}_i^{t}$} denotes the {\footnotesize$i$}-th input image, and {\footnotesize$L_i^{t}$} denotes its detection annotations. Specifically,
{\footnotesize $L_i^{t}=\{(\mathbf{b}_{i,k}^{t},\mathbf{c}_{i,k}^{t})\}_{k=1}^{K_i}$},
where {\footnotesize$K_i$} is the number of annotated objects in {\footnotesize $\mathbf{I}_i^{t}$}, {\footnotesize $\mathbf{b}_{i,k}^{t}$} denotes the bounding box of the {\footnotesize$k$}-th object, and {\footnotesize $\mathbf{c}_{i,k}^{t}$} denotes its concept label, with {\footnotesize $\mathbf{c}_{i,k}^{t}\in\mathcal{C}^{t}$}.
At each task, the model can access only the current-task data {\footnotesize $\mathcal{D}^{t}$}, while image data from all previous tasks are unavailable.

During evaluation, the updated detector {\footnotesize $f_{\theta_t}$} is required to detect objects from three sets: the current novel concepts {\footnotesize $\mathcal{C}^{t}$}, the previously learned concepts
{\footnotesize $\mathcal{C}^{\mathrm{his},t}=\textstyle\bigcup_{k=1}^{t-1}\mathcal{C}^{k}$},
and the original open-vocabulary concept set {\footnotesize $\mathcal{C}^{\mathrm{ovd}}$}, where {\footnotesize$\mathcal{C}^{\mathrm{his},t}$} denotes all novel categories introduced before task {\footnotesize $t$}, and {\footnotesize $\mathcal{C}^{\mathrm{ovd}}$} denotes the concepts space supported by the initial detector.
Therefore, COVD requires the model to continually incorporate novel concepts while preserving detection performance on both previously learned concepts and the original open-vocabulary categories.

\subsection{Overview of the Method} 
\begin{figure*}[t!]
	\centering
	\includegraphics[width=0.98\linewidth]{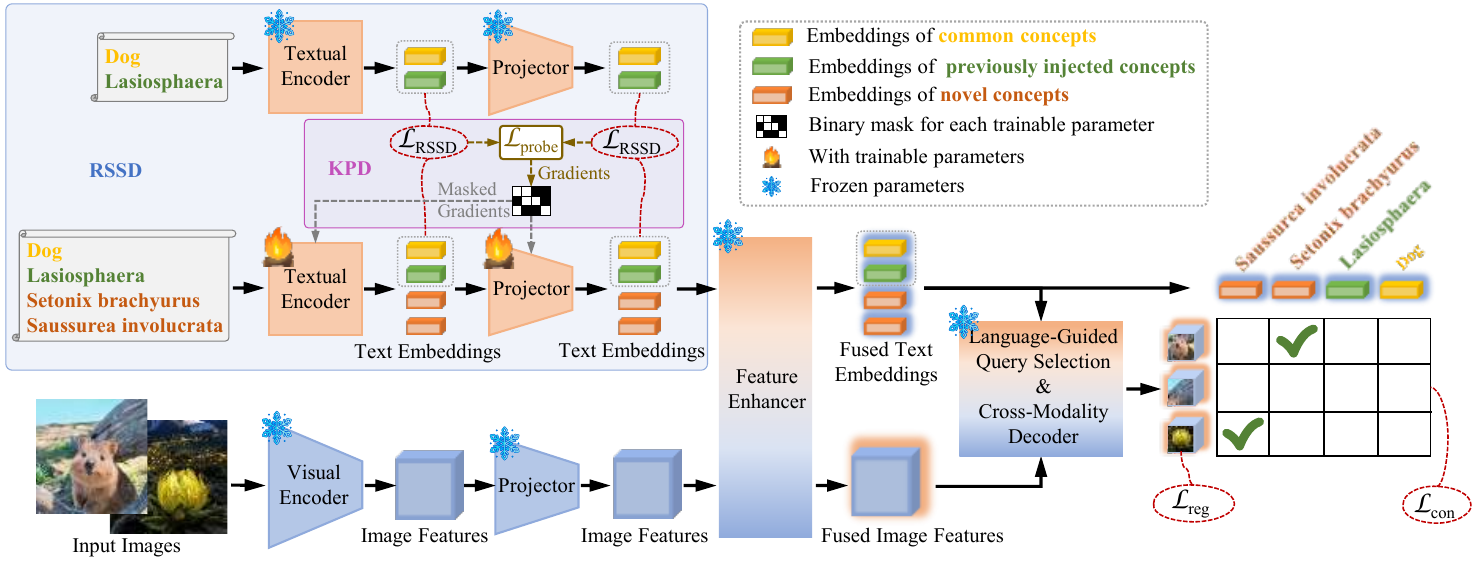}  \vspace{-10pt}
    \caption{Overview of NoIn-Det. Built on LLMDet, NoIn-Det freezes the visual encoder and updates only a small subset of textual parameters for continual novel concept injection. At each task, it learns current concepts via region-text contrastive learning, while randomly sampled common concepts and stored historical concepts provide lightweight textual knowledge for stability regularization. \textbf{Representation Space Stability Distillation (RSSD)} preserves representation stability, and \textbf{Knowledge-aware Parameter Decoupling (KPD)} guides parameter selection, enabling continual concept learning without extra parameters or historical image replay.}  
	\label{fig:overview}
	\vspace{-15pt}
\end{figure*}

Our goal is to continually inject novel concepts into an open-vocabulary detector LLMDet in a parameter-efficient and stable manner. As shown in Fig.~\ref{fig:overview}, NoIn-Det freezes the visual encoder of a pre-trained open-vocabulary detector and updates only a small subset of key textual parameters. For each task, it learns current concepts via contrastive learning on region-text pairs, while using sampled common concepts and stored historical concept texts as lightweight old-knowledge supervision. We introduce \textbf{Representation Space Stability Distillation (RSSD)} to preserve semantic consistency and structural stability in the textual embedding space, and \textbf{Knowledge-aware Parameter Decoupling (KPD)} to suppress drift in old-knowledge-sensitive parameters. Thus, NoIn-Det enables continual novel concept injection without additional parameters or historical image replay.

\subsection{Representation Space Stability Distillation (RSSD)}  
In continual novel concept learning, new-task training can shift the textual representation space, causing old concept forgetting and weakening the open-vocabulary foundation.
To address this, we propose \textbf{RSSD}, which preserves textual-space stability via lightweight stored textual knowledge distillation without historical images.

\textbf{\textit{Lightweight textual old knowledge.}}
At task $t$, we construct an old-knowledge concept set
{\footnotesize $\mathcal{C}_{\mathrm{old}}^{t}=\mathcal{C}_{\mathrm{com}}^{t}\cup\mathcal{C}_{\mathrm{his}}^{t}$},
where {\footnotesize $\mathcal{C}_{\mathrm{com}}^{t}$} denotes a randomly sampled set of common concepts, and
{\footnotesize $\mathcal{C}_{\mathrm{his}}^{t}=\bigcup_{k=1}^{t-1}\mathcal{C}^{k}$}
denotes the set of all concepts learned before task $t$. Here, {\footnotesize $\mathcal{C}^{k}$} is the set of novel concepts introduced at task $k$. We further denote the current novel concept set by
{\footnotesize $\mathcal{C}_{\mathrm{novel}}^{t}=\mathcal{C}^{t}$}.
Accordingly, the student text branch receives
{\footnotesize $\mathcal{C}_{\mathrm{stu}}^{t}=\mathcal{C}_{\mathrm{old}}^{t}\cup\mathcal{C}_{\mathrm{novel}}^{t}$},
while the teacher text branch only takes the old-knowledge texts:
{\footnotesize $\mathcal{C}_{\mathrm{tea}}^{t}=\mathcal{C}_{\mathrm{old}}^{t}$}.
Since common concepts are easy to obtain and historical concept are cheap to store, this design provides effective old-knowledge supervision without any historical image replay.

\textbf{\textit{Teacher-student textual representations.}}
We use the text branch (frozen) from the previous task as the teacher and the current text branch (trainable) as the student. Let {\footnotesize $E_{\mathrm{tea}}(\cdot)$} and {\footnotesize $P_{\mathrm{tea}}(\cdot)$} denote the teacher textual encoder and projector, respectively, and let {\footnotesize $E_{\mathrm{stu}}(\cdot)$} and {\footnotesize $P_{\mathrm{stu}}(\cdot)$} denote the corresponding student modules.
For the teacher and student branches, old-knowledge texts are encoded as
\begin{equation}
\footnotesize
\begin{aligned}
\mathbf{E}_{\mathrm{tea}}^{t}&=E_{\mathrm{tea}}(\mathcal{C}_{\mathrm{old}}^{t}), &
\mathbf{P}_{\mathrm{tea}}^{t}&=P_{\mathrm{tea}}(\mathbf{E}_{\mathrm{tea}}^{t}), &
\mathbf{E}_{\mathrm{stu,old}}^{t}&=E_{\mathrm{stu}}(\mathcal{C}_{\mathrm{old}}^{t}), &
\mathbf{P}_{\mathrm{stu,old}}^{t}&=P_{\mathrm{stu}}(\mathbf{E}_{\mathrm{stu,old}}^{t}),
\end{aligned}
\end{equation}
where {\footnotesize $\mathbf{E}_{\mathrm{tea}}^{t}$} and {\footnotesize $\mathbf{E}_{\mathrm{stu,old}}^{t}$} denote the teacher and student encoder-level text embeddings, respectively, and {\footnotesize $\mathbf{P}_{\mathrm{tea}}^{t}$} and {\footnotesize $\mathbf{P}_{\mathrm{stu,old}}^{t}$} denote their projected embeddings for old-knowledge concepts.

\textbf{\textit{Dual distillation for representation space stability.}}
RSSD preserves textual representation stability by imposing two complementary constraints on old-knowledge texts: absolute semantic distillation and relative relation distillation.
The absolute semantic distillation aligns student and teacher representations of the same old concepts at both encoder and projector levels:
\begin{equation}
\footnotesize
\mathcal{L}_{\mathrm{abs}}^{E,t}
=
1-\mathrm{Cos}\!\left(\mathbf{E}_{\mathrm{stu,old}}^{t},\mathbf{E}_{\mathrm{tea}}^{t}\right), \qquad
\mathcal{L}_{\mathrm{abs}}^{P,t}
=
1-\mathrm{Cos}\!\left(\mathbf{P}_{\mathrm{stu,old}}^{t},\mathbf{P}_{\mathrm{tea}}^{t}\right),
\end{equation}
where {\footnotesize $\mathrm{Cos}(\cdot,\cdot)$} denotes average cosine similarity between corresponding student and teacher features.

To preserve old textual-space structure, we impose relative relation distillation. For a feature set {\footnotesize $\mathbf{X}=[\mathbf{x}_1,\mathbf{x}_2,\dots,\mathbf{x}_N]^\top$}, its pairwise semantic relation matrix is defined as
{\footnotesize $\mathbf{R}(\mathbf{X})=
\left[
\frac{\mathbf{x}_i^\top \mathbf{x}_j}{\|\mathbf{x}_i\|_2 \, \|\mathbf{x}_j\|_2}
\right]_{i,j=1}^{N}$},
where {\footnotesize $N$} is the number of concept embeddings in {\footnotesize $\mathbf{X}$}, and $i,j$ index corresponding embeddings.

Accordingly, the teacher and student relation matrices are computed as
\begin{equation}
\footnotesize
\begin{aligned}
\mathbf{R}_{\mathrm{tea}}^{E,t}&=\mathbf{R}(\mathbf{E}_{\mathrm{tea}}^{t}), &
\mathbf{R}_{\mathrm{stu}}^{E,t}&=\mathbf{R}(\mathbf{E}_{\mathrm{stu,old}}^{t}), &
\mathbf{R}_{\mathrm{tea}}^{P,t}&=\mathbf{R}(\mathbf{P}_{\mathrm{tea}}^{t}), &
\mathbf{R}_{\mathrm{stu}}^{P,t}&=\mathbf{R}(\mathbf{P}_{\mathrm{stu,old}}^{t}).
\end{aligned}
\end{equation}
Then, the relation distillation losses are defined as
\begin{equation}
\footnotesize
\mathcal{L}_{\mathrm{rel}}^{E,t}
=
\mathrm{SmoothL1}\!\left(\mathbf{R}_{\mathrm{stu}}^{E,t},\mathbf{R}_{\mathrm{tea}}^{E,t}\right), \qquad
\mathcal{L}_{\mathrm{rel}}^{P,t}
=
\mathrm{SmoothL1}\!\left(\mathbf{R}_{\mathrm{stu}}^{P,t},\mathbf{R}_{\mathrm{tea}}^{P,t}\right).
\end{equation}
Combining the above terms, the overall RSSD loss is written as
{\footnotesize 
$ \mathcal{L}_{\mathrm{RSSD}}^{t}
=
\mathcal{L}_{\mathrm{abs}}^{E,t}
+\mathcal{L}_{\mathrm{abs}}^{P,t}
+\mathcal{L}_{\mathrm{rel}}^{E,t}
+\mathcal{L}_{\mathrm{rel}}^{P,t}$}.

\textbf{\textit{Joint optimization with current-task learning.}}
At task {\footnotesize $t$}, the visual branch only receives images of current novel concepts, denoted as {\footnotesize $\mathcal{I}^{t}$}, while the student text branch contains both old and new concept texts.
Thus, the student old-knowledge text embeddings are not only regularized by RSSD, but also serve as negatives in current cross-modal contrastive learning with image features from {\footnotesize $\mathcal{I}^{t}$}. Let {\footnotesize $\mathcal{L}_{\mathrm{reg}}^{t}$} and {\footnotesize $\mathcal{L}_{\mathrm{con}}^{t}$} denote the detection loss and cross-modal contrastive loss at task {\footnotesize $t$}, respectively. The overall objective is
{\footnotesize 
$\mathcal{L}_{\mathrm{task}}^{t}
=\mathcal{L}_{\mathrm{reg}}^{t}
+\mathcal{L}_{\mathrm{con}}^{t}
+\mathcal{L}_{\mathrm{RSSD}}^{t}$}.
In this way, RSSD preserves both individual old-concept semantics and textual-space structure, while anchoring newly learned concepts in a stable semantic space without storing historical images.

\subsection{Knowledge-aware Parameter Decoupling (KPD)}  
Although RSSD preserves old knowledge at the representation level, overly strong constraints may hinder novel concept injection. We argue that not all trainable parameters should equally serve old-knowledge retention. To balance retention and injection, we further propose \textbf{KPD}, which uses old-knowledge distillation signals to identify old-knowledge-sensitive parameters, constrain their updates, and free the remaining parameters to learn novel concepts.

\textbf{\textit{Restricted trainable parameter subspace.}}
To maintain parameter efficiency, we optimize only a small text-side parameter subset, including the last several layers of the textual encoder and the projector, while freezing the rest of the detector. Formally, let {\footnotesize $\Theta$} be the full detector parameter set and {\footnotesize $\Theta_{\mathrm{tr}} \subset \Theta$} the trainable subset for continual learning. This confines novel concept incorporation to a compact parameter subspace, thereby reducing the risk of overwriting old knowledge.

\textbf{\textit{Old-knowledge-sensitive probe gradients.}}
At task {\footnotesize $t$}, we use the preserved old-knowledge concept set {\footnotesize $\mathcal{C}_{\mathrm{old}}^{t}$} to estimate the sensitivity of trainable parameters to old knowledge via a lightweight probe loss defined only on old-knowledge texts. Using the same teacher-student textual representations as in RSSD, we define
{\footnotesize $\mathcal{L}_{\mathrm{probe}}^{t}=\mathcal{L}_{\mathrm{abs}}^{E,t}+\mathcal{L}_{\mathrm{abs}}^{P,t}$},
this probe loss is used only to estimate parameter sensitivity to old knowledge, rather than serving as the final training objective.
For each trainable parameter element $\theta_i \in \Theta_{\mathrm{tr}}$, we compute its probe gradient as
{\footnotesize $g_i^{\mathrm{probe}}=\frac{\partial \mathcal{L}_{\mathrm{probe}}^{t}}{\partial \theta_i}$},
and the magnitude {\footnotesize$\left|g_i^{\mathrm{probe}}\right|$} indicates how sensitive {\footnotesize$\theta_i$} is to the preserved old knowledge: a larger magnitude suggests that changing this element is more likely to disturb previously acquired semantics.

\textbf{\textit{Element-wise gradient masking.}}
Based on the probe gradients, we construct an element-wise gradient mask to suppress updates on old-knowledge-sensitive parameters. Specifically, we collect the absolute probe gradients of all trainable parameter elements in {\footnotesize$\Theta_{\mathrm{tr}}$} and compute a quantile threshold
\begin{equation}
\footnotesize
\label{eq:Grad}
\tau^{t}=Q_{r}\!\left(\left\{\left|g_i^{\mathrm{probe}}\right| \mid \theta_i \in \Theta_{\mathrm{tr}}\right\}\right),
\end{equation}
where {\footnotesize$Q_r(\cdot)$} denotes the {\footnotesize$r$}-th quantile operator, and {\footnotesize$r\in(0,1)$} controls the proportion of retained trainable parameter elements.
We then define the binary mask for each trainable parameter element as
{\footnotesize$m_i^{t}=\mathbf{1}\!\left(\left|g_i^{\mathrm{probe}}\right|\le \tau^{t}\right)$},
where {\footnotesize$\mathbf{1}(\cdot)$} is the indicator function. Thus, parameter elements with relatively small probe gradients are retained for updating, while those with large probe gradients are suppressed.
Let $g_i^{\mathrm{train}}$ denote the training-loss gradient w.r.t. {\footnotesize$\theta_i$}. KPD modifies it by
{\footnotesize
$\tilde{g}_i^{\mathrm{train}}
=
m_i^{t}\odot g_i^{\mathrm{train}}$},
where {\footnotesize$\odot$} denotes element-wise multiplication. The effective parameter update is then performed using {\footnotesize$\tilde{g}_i^{\mathrm{train}}$} instead of {\footnotesize$g_i^{\mathrm{train}}$}. In this way, KPD filters out high-risk updates that are likely to harm old knowledge, while leaving a safer parameter subspace for continual novel concept learning.

\textbf{\textit{KPD in training.}}
KPD does not replace the original objective, but acts as an update constraint complementary to RSSD and current-task supervision. During back-propagation, KPD operates on updates of {\footnotesize$\Theta_{\mathrm{tr}}$} to \emph{identify and suppress old-knowledge-sensitive parameters}, releasing more parameter space for novel concept learning. This enables better adaptation to current novel concepts while avoiding destructive changes to existing semantic knowledge. Overall, RSSD stabilizes the textual semantic space at the representation level, while KPD decouples retention and learning at the parameter level, enabling more stable continual novel concept incorporation.

\textbf{Implementation Details.}
All experiments are conducted on 8 NVIDIA 3090 GPUs with a total batch size of 16. The model is trained for 24 epochs using AdamW with an initial learning rate of 1e-3 and weight decay of 1e-4, decayed by 0.1 at epochs 11 and 20.
We randomly sample 100 common concepts from ImageNet~\cite{deng2009imagenet} and Objects365~\cite{shao2019objects365}. The {\footnotesize $r$} in Eq.~(\ref{eq:Grad}) is set to 0.5.
\textit{At the beginning of task {\footnotesize $t$}, the model is initialized from task {\footnotesize $t\!-\!1$}; the student and teacher textual branches share this initialization, after which the teacher is frozen and the student is updated during training.}

\section{Experiments}
\subsection{Setup}

\textbf{COVD settings.} 
We build the continual novel concept incorporation benchmark on Novel-114. As described in Sec.~\ref{sec:Benchmark}, Novel-114 is divided into seven sequential tasks: Animals-25, Clothing-6, Cold Weapons-4, Food-28, Instruments-8, Plants-25, and Vessels-18. Unless specified, all experiments follow this order.
The model incorporates concepts task by task. At task {\footnotesize$t$}, it can access only current-task training images and annotations, without historical images. This setting evaluates novel concept learning and old knowledge retention without image replay.
After each task, we evaluate the model on the current and all previous tasks to measure acquisition and retention, and on MS COCO to examine whether original open-vocabulary generalization is preserved.

\textbf{Comparison methods.} 
Since this paper studies a new task, continual concept injection for OVD, and introduces a new benchmark, there is no directly applicable standard baseline in existing literature. We therefore select baselines from the perspectives of task relevance and method representativeness, including ZiRa~\cite{deng2024zero} and SGVF~\cite{cheng2026incremental}, which are more related to detection-style VLMs, and MoE-Adapters~\cite{yu2024boosting}, LADA~\cite{luo2025lada}, C-CLIP~\cite{liu2025c}, and KeepLoRA~\cite{luo2026keeplora}, which represent adapter, MoE, reparameterization, and LoRA paradigms in continual learning for classification VLMs.
For fair comparison, we adopt LLMDet-Tiny as the unified base model, adapt each method to the LLMDet~\cite{fu2025llmdet} framework while preserving its core design as much as possible, and compare them under the same training configuration and evaluation protocol.

\textbf{Evaluation protocol and metrics.}
To evaluate the stability--plasticity trade-off in continual novel concept injection, we report results from three perspectives: \textbf{\ding{172} \textit{Original knowledge retention}}, \textbf{\ding{173} \textit{Novel concept acquisition}}, and \textbf{\ding{174} \textit{Historical injected concept retention}}.
The \graycell{gray row} in Table~\ref{tab:results} reports the zero-shot performance of LLMDet before injection. Its low performance on most Novel-114 tasks shows that these categories cannot be well detected using existing open-vocabulary knowledge alone, validating the need for Novel-114.
For original knowledge retention, we evaluate MS COCO~\citep{lin2014microsoft} after each stage, with \redcell{red cells} indicating preserved OVD capability. For novel concept acquisition, we test the current task after training, with \yellowcell{yellow diagonal cells} showing adaptation to novel concepts. For historical concept retention, we evaluate all previous tasks after each subsequent stage, with white cells reporting retention performance. Finally, the \bluecell{blue cells} summarize the final average performance after all seven tasks, jointly reflecting acquisition, retention, and original OVD preservation.
All experiments use COCO-style detection metrics, $AP_{50:95}$ and $AP_{50}$, with each entry reported as $AP_{50:95}$/$AP_{50}$.

\begin{table*}[t!]
\centering
\tiny
\caption{The main results on MS COCO and Novel-114 (\%).}
\vspace{2pt}
\label{tab:results}
\resizebox{\textwidth}{!}{
\begin{tabular}{l|l|c|c|c|c|c|c|c|c|c}
\hline
\multirow{2}{*}{Method} & \multirow{2}{*}{\diagbox[width=8em]{Train}{Test}} & \multirow{2}{*}{MS COCO} & \multirow{2}{*}{Animals-25} & \multirow{2}{*}{Clothing-6} & \multirow{2}{*}{Cold Weapons-4} & \multirow{2}{*}{Food-28} & \multirow{2}{*}{Instruments-8} & \multirow{2}{*}{Plants-25} & \multirow{2}{*}{Vessels-18} & \multirow{2}{*}{Avg.} \\
 &  & &  &  &  &  &  &  &  & \\
\hline
\rowcolor{gray!15} LLMDet &--  & 54.4/70.8 & 5.9/7.3   & 10.7/22.4 & 27.1/32.0 & 9.5/13.5  & 31.3/32.7 & 3.5/5.1   & 4.1/4.8   & 18.3/23.6 \\
\hline

\multirow{7}{*}{ZiRa}
& Animals-25      &\cellcolor{red!10} 46.4/60.7 &\cellcolor{yellow!20} 86.6/94.5 & --        & --        & --        & --        & --        & --        & --        \\
& Clothing-6      &\cellcolor{red!10} 43.7/57.2 &85.2/93.2  & \cellcolor{yellow!20}43.9/49.1 & --        & --        & --        & --        & --        & --        \\
& Cold Weapons-4  &\cellcolor{red!10}43.0/56.4  &69.1/75.5  &37.0/40.5  & \cellcolor{yellow!20}30.0/36.0 & --        & --        & --        & --        & --        \\
& Food-28         &\cellcolor{red!10}34.5/45.5  &58.9/66.3  &14.8/22.7   &20.2/24.8  & \cellcolor{yellow!20}70.5/80.7 & --        & --        & --        & --        \\
& Instruments-8   &\cellcolor{red!10}35.8/47.2  &61.6/69.1  &13.8/20.0  &18.6/23.3  &68.7/78.8  & \cellcolor{yellow!20}44.0/51.0 & --        & --        & --        \\
& Plants-25       &\cellcolor{red!10}19.7/26.4  &27.0/30.4  &11.3/17.6   &29.2/34.0  &1.3/1.5  &13.5/15.0  & \cellcolor{yellow!20}66.3/85.8 & --        & --        \\
& Vessels-18      &\cellcolor{red!10}17.7/23.2  &31.3/35.1   & 11.8/15.8  &20.8/24.1   &38.2/45.8  &21.5/24.3  &61.0/78.8  & \cellcolor{yellow!20} 55.9/59.4 & \cellcolor{blue!15}32.3/38.3  \\
\hline
\multirow{7}{*}{MoE-Adapters}
& Animals-25      &\cellcolor{red!10} 39.4/52.3 & \cellcolor{yellow!20}84.5/91.6 & --        & --        & --        & --        & --        & --        & -- \\
& Clothing-6      &\cellcolor{red!10} 36.4/48.7 & 64.9/70.7 & \cellcolor{yellow!20}33.5/37.4 & --        & --        & --        & --        & --        & -- \\
& Cold Weapons-4  &\cellcolor{red!10} 31.3/42.5 & 58.6/64.6 & 32.0/34.0 & \cellcolor{yellow!20}59.0/64.8 & --        & --        & --        & --        & -- \\
& Food-28         &\cellcolor{red!10} 31.3/42.5 & 65.4/72.3 & 25.7/29.0 & 34.7/40.4 & \cellcolor{yellow!20}64.8/74.2 & --        & --        & --        & -- \\
& Instruments-8   &\cellcolor{red!10} 38.1/52.0 & 49.5/55.1 & 28.0/31.2 & 30.6/35.4 & 62.5/72.0 & \cellcolor{yellow!20}31.1/32.2 & --        & --        & -- \\
& Plants-25       &\cellcolor{red!10} 33.5/45.7 & 56.0/63.2 & 27.2/33.0 & 20.7/23.9 & 48.7/57.5 & 12.4/17.6 & \cellcolor{yellow!20}42.7/55.0 & --        & -- \\
& Vessels-18      &\cellcolor{red!10} 30.7/42.0 & 58.4/65.1 & 28.8/36.8 & 23.2/26.8 & 47.1/55.4 & 16.2/19.8 & 38.5/50.0 & \cellcolor{yellow!20}60.3/61.3 & \cellcolor{blue!15} 37.9/44.7 \\
\hline
\multirow{7}{*}{LADA}
& Animals-25      &\cellcolor{red!10} 35.7/47.2 & \cellcolor{yellow!20}87.9/95.0 & --        & --        & --        & --        & --        & --        & -- \\
& Clothing-6      &\cellcolor{red!10} 17.6/22.6 & 69.7/80.9 & \cellcolor{yellow!20}64.2/68.2 & --        & --        & --        & --        & --        & -- \\
& Cold Weapons-4  &\cellcolor{red!10} 20.2/26.7 & 63.6/71.1 & 39.7/44.9 & \cellcolor{yellow!20}75.3/81.8 & --        & --        & --        & --        & -- \\
& Food-28         &\cellcolor{red!10} 32.4/43.8 & 34.8/41.1 & 13.8/19.3 & 26.1/28.7 & \cellcolor{yellow!20}74.3/84.9 & --        & --        & --        & -- \\
& Instruments-8   &\cellcolor{red!10} 18.1/24.4 & 15.4/20.6 & 17.9/23.5 & 23.9/28.1 & 56.5/66.7 & \cellcolor{yellow!20}57.9/61.9 & --        & --        & -- \\
& Plants-25       &\cellcolor{red!10} 23.4/31.8 & 20.2/25.0 & 5.6/10.6  & 20.2/25.0 & 27.5/35.0 & 34.2/38.9 & \cellcolor{yellow!20}70.1/89.3 & --        & -- \\
& Vessels-18      &\cellcolor{red!10} 20.8/27.9 & 16.0/20.5 & 6.0/10.6    & 20.6/23.6 & 9.3/13.1  & 38.4/52.6 & 12.3/14.1 & \cellcolor{yellow!20}70.9/71.9 & \cellcolor{blue!15} 24.3/29.3 \\
\hline
\multirow{7}{*}{C-CLIP}
& Animals-25      &\cellcolor{red!10} 48.3/64.0 & \cellcolor{yellow!20}64.4/71.7 & --        & --        & --        & --        & --        & --        & -- \\
& Clothing-6      &\cellcolor{red!10} 43.3/58.0 & 59.0/66.0 & \cellcolor{yellow!20}42.0/45.7 & --        & --        & --        & --        & --        & -- \\
& Cold Weapons-4  &\cellcolor{red!10} 45.2/61.5 & 27.6/32.5 & 30.4/32.8 & \cellcolor{yellow!20}51.4/60.6 & --        & --        & --        & --        & -- \\
& Food-28         &\cellcolor{red!10} 45.4/63.1 & 17.6/20.9 & 11.3/19.3 & 34.2/40.4 & \cellcolor{yellow!20}46.8/55.9 & --        & --        & --        & -- \\
& Instruments-8   & \cellcolor{red!10}46.6/62.6 & 14.8/17.1 & 17.9/26.3 & 30.7/35.3 & 31.5/37.4 & \cellcolor{yellow!20}41.9/45.5 & --        & --        & -- \\
& Plants-25       &\cellcolor{red!10} 32.7/45.4 & 9.0/11.1  & 17.4/25.4 & 26.8/31.1 & 12.0/14.7 & 12.5/17.5 & \cellcolor{yellow!20}32.6/43.0 & --        & -- \\
& Vessels-18      &\cellcolor{red!10} 32.7/45.6 & 4.4/5.9   & 16.3/24.2 & 25.7/31.6 & 7.1/10.0  & 10.8/16.5 & 18.6/24.3 & \cellcolor{yellow!20}23.7/26.3 & \cellcolor{blue!15}17.4/23.1 \\
\hline
\multirow{7}{*}{SGVF}
& Animals-25      &\cellcolor{red!10} 26.6/35.9 & \cellcolor{yellow!20}83.7/92.0 & --        & --        & --        & --        & --        & --        & -- \\
& Clothing-6      &\cellcolor{red!10} 29.5/39.1 & 82.2/91.6 & \cellcolor{yellow!20}61.1/67.4 & --        & --        & --        & --        & --        & -- \\
& Cold Weapons-4  &\cellcolor{red!10} 26.3/35.1 & 78.8/88.4 & 39.1/45.6 & \cellcolor{yellow!20}73.4/79.3 & --        & --        & --        & --        & -- \\
& Food-28         &\cellcolor{red!10} 24.7/33.7 & 60.1/76.5 & 24.3/31.9 & 44.8/51.0 & \cellcolor{yellow!20}71.2/82.0 & --        & --        & --        & -- \\
& Instruments-8   &\cellcolor{red!10} 24.0/32.6 & 52.0/70.5 & 16.9/20.4 & 40.0/46.8 & 64.3/74.1 & \cellcolor{yellow!20}64.2/68.6 & --        & --        & -- \\
& Plants-25       &\cellcolor{red!10} 24.7/33.7 & 59.8/70.9 & 16.5/20.4 & 46.8/41.1 & 52.3/62.7 & 47.2/50.4 & \cellcolor{yellow!20}65.9/85.9 & --        & -- \\
& Vessels-18      &\cellcolor{red!10} 19.3/26.4 & 52.1/63.1 & 11.9/15.1 & 39.4/47.0 & 42.7/52.0 & 36.3/39.1 & 39.3/54.5 & \cellcolor{yellow!20}73.4/74.7 & \cellcolor{blue!15} 39.3/46.5 \\
\hline
\multirow{7}{*}{KeepLoRa}
& Animals-25      &\cellcolor{red!10} 50.2/67.6 & \cellcolor{yellow!20}14.9/19.8 & --        & --        & --        & --        & --        & --        & -- \\
& Clothing-6      &\cellcolor{red!10} 29.5/40.6 & 2.6/3.8 & \cellcolor{yellow!20}8.7/17.7 & --        & --        & --        & --        & --        & -- \\
& Cold Weapons-4  &\cellcolor{red!10} 34.5/47.1 & 5.4/7.0 & 9.8/22.5 & \cellcolor{yellow!20}26.7/32.0 & --        & --        & --        & --        & -- \\
& Food-28         &\cellcolor{red!10} 45.1/60.3 & 7.1/8.8 & 6.7/18.1 & 29.6/36.8 & \cellcolor{yellow!20}17.3/23.6 & --        & --        & --        & -- \\
& Instruments-8   &\cellcolor{red!10} 31.9/46.2 & 4.9/6.2 & 9.4/17.9  & 21.0/25.9 & 6.9/10.4  & \cellcolor{yellow!20}23.8/26.8  & --        & --        & -- \\
& Plants-25       &\cellcolor{red!10} 44.7/59.9 & 7.0/9.2 & 13.2/29.0 & 26.5/32.7 & 9.3/14.0  & 26.7/30.4 & \cellcolor{yellow!20}5.5/8.0 & --        & -- \\
& Vessels-18      &\cellcolor{red!10} 43.8/59.7 & 6.1/7.6 & 10.1/18.8  & 25.2/31.0  & 8.4/13.3 & 28.2/32.1 & 3.7/5.2  & \cellcolor{yellow!20}6.8/8.9 & \cellcolor{blue!15}16.5/22.1 \\
\hline
\multirow{7}{*}{\begin{tabular}[l]{@{}l@{}}
NoIn-Det \\
\end{tabular}}
& Animals-25     &\cellcolor{red!10} 49.8/65.1 &\cellcolor{yellow!20}87.9/95.2 & --        & --        & --        & --        & --        & --        & --        \\
& Clothing-6     &\cellcolor{red!10} 49.5/64.3  &80.7/88.6  & \cellcolor{yellow!20}47.1/54.6 & --        & --        & --        & --        & --        & --        \\
& Cold Weapons-4 &\cellcolor{red!10} 49.1/63.7  &81.9/89.0  & 40.5/53.5 &\cellcolor{yellow!20}71.0/79.6  & --        & --        & --        & --        & --        \\
& Food-28        &\cellcolor{red!10} 47.0/62.0 &73.3/81.0   &24.3/32.7  &55.8/62.7  &\cellcolor{yellow!20}72.8/82.9 &-- &-- &-- & -- \\
& Instruments-8  &\cellcolor{red!10}47.1/61.9  &75.9/83.4  &14.9/25.0  &51.0/61.3  &68.8/78.7  &\cellcolor{yellow!20}62.5/67.1 & --        & --        & --        \\
& Plants-25      &\cellcolor{red!10}45.7/60.2  &48.7/54.5  &20.6/29.6  &44.7/50.1  &55.3/64.8  &63.3/69.5   &\cellcolor{yellow!20}69.6/89.1 & --        & --        \\
& Vessels-18     &\cellcolor{red!10}41.7/54.8  &50.1/57.3  & 18.2/26.7 &31.1/38.8  &43.2/54.1  &35.7/39.2  &63.7/83.3 &\cellcolor{yellow!20}70.4/73.7 & \cellcolor{blue!15} 44.3/53.5 \\
\hline
\end{tabular}}\vspace{-20pt}
\end{table*}

\subsection{Main Results}
Table~\ref{tab:results} compares our method with existing approaches on MS COCO and Novel-114 under continual novel concept injection.
Overall, our method achieves superior performance across most tasks and protocols. Compared with prior SOTA methods such as SGVF and MoE-Adapters, our method obtains the best final average performance (44.3/53.5), demonstrating strong capability in balancing stability and plasticity during continual learning.
\textbf{\textit{\ding{172} Original knowledge retention.}} 
From the red-highlighted regions, which reflect the performance on MS COCO after each stage of novel concept injection, our method maintains significantly higher accuracy compared to other methods. This indicates that the proposed approach effectively mitigates catastrophic forgetting and preserves the original open-vocabulary detection (OVD) capability.
\textbf{\textit{\ding{173} Novel concept acquisition.}} 
From the yellow diagonal entries, which correspond to the performance on the current task after each training stage, our method achieves competitive or superior results across all tasks. This demonstrates that the model retains strong plasticity and is able to rapidly adapt to newly introduced concepts without sacrificing learning efficiency.
\textbf{\textit{\ding{174} Historical injected concept retention.}} 
From the off-diagonal white regions, our method shows more stable performance on previously learned tasks compared to baseline methods. This suggests that the proposed design enables better knowledge consolidation, reducing the degradation of earlier learned concepts during subsequent training stages.
\textbf{\textit{Overall performance.}} 
Finally, the blue-highlighted results show that our method achieves the highest average performance among all approaches, confirming that it effectively balances preserving original knowledge, learning new concepts, and retaining historical knowledge.
In contrast, KeepLoRA and C-CLIP suffer from severe forgetting or insufficient adaptation, while SGVF shows strong plasticity but weaker stability. Our method achieves a better trade-off, leading to consistent improvements across evaluation metrics.

\subsection{Ablation Study}
\textbf{Component ablation.}
As shown in Table~\ref{tab:components}, we conduct ablation studies on MS COCO and Novel-114 to evaluate the contributions of RSSD and KPD in NoIn-Det. Since KPD relies on the stable gradient signals induced by RSSD for parameter-level constraints, it cannot be applied independently. We therefore adopt a progressive ablation strategy: adding RSSD first to stabilize the representation space, and then adding KPD to further constrain parameter updates.

\textbf{\textit{Effect of RSSD.}}
Compared with LLMDet, LLMDet+RSSD improves original OVD knowledge retention. As shown by the red-highlighted results, the model retains higher COCO performance after continual learning, indicating that RSSD mitigates forgetting by stabilizing the textual representation space. Meanwhile, strong current-task performance shows that RSSD does not hinder plasticity.

\textbf{\textit{Effect of KPD.}}
Adding KPD brings further gains. COCO retention improves from 39.0\%/52.4\% to 47.1\%/61.7\%, and final average performance increases from 40.4\%/49.3\% to 44.3\%/53.5\%. This shows that KPD strengthens knowledge consolidation while preserving novel concept learning.

\textbf{\textit{Overall analysis.}}
RSSD and KPD are complementary: RSSD stabilizes the semantic representation space, while KPD regularizes parameter updates at a finer granularity. Together, they achieve a better stability--plasticity trade-off and superior overall performance.

\textbf{Effect of fine-tuned layers in the textual encoder.}
We further study how the number of fine-tuned top layers in the textual encoder affects RSSD, as shown in Table~\ref{tab:layer_ablation}. We vary the number of updated layers from 0 to 12 to analyze the stability--plasticity trade-off.
Updating no layers best preserves original knowledge (52.7\%/68.6\%) but provides limited adaptation. As more layers are fine-tuned, novel concept acquisition improves, especially with 4--10 updated layers, but OVD retention gradually decreases, indicating stronger forgetting. Updating too few layers limits learning, while updating too many layers disrupts the original semantic space.
The best final performance is achieved by fine-tuning a moderate number of layers, \textit{e.g.}, 4 layers, which obtains the highest average score (44.3\%/53.5\%). This suggests that lower textual layers encode more general and transferable semantics and should remain stable, whereas higher layers are more task-adaptive. Selectively updating top layers therefore enables effective concept injection while preserving the core representation space.

\textbf{Effect of update ratio in KPD.}
We analyze the sensitivity of the update ratio $r$ in KPD, which controls the proportion of trainable parameter elements to update, as shown in Table~\ref{tab:r_sensitivity}.
As $r$ increases, the model generally gains stronger novel concept acquisition, since more parameters are allowed to adapt. However, this also leads to more severe OVD forgetting. When $r=1.0$, the model suffers the largest degradation, with the lowest COCO retention (39.0\%/52.4\%), while smaller ratios better preserve original knowledge. This shows that unrestricted parameter updates can damage the original open-vocabulary capability.
The final results reveal a clear stability--plasticity trade-off. A too small $r$ limits adaptation, whereas a too large $r$ causes excessive forgetting. The best performance is achieved at $r=0.5$, obtaining the highest average score (44.3\%/53.5\%). These results confirm that KPD benefits from selective parameter updates, enabling the model to incorporate novel concepts while preserving previously learned representations.
\textit{\textbf{Additional ablations, second-round concept injection, and qualitative results are provided in Appendix Sections~\ref{sec:Common_Concept_Sampling}, \ref{sec:Second}, and \ref{sec:Qualitative}, respectively.}}

\begin{table*}[t!]
\centering
\tiny
\caption{Component ablation of RSSD and KPD on MS COCO and Novel-114 (\%).}
\label{tab:components}
\resizebox{\textwidth}{!}{
\begin{tabular}{l|l|c|c|c|c|c|c|c|c|c}
\hline
Method & Protocol & MS COCO & Animals-25 & Clothing-6 & Cold Weapons-4 & Food-28 & Instruments-8 & Plants-25 & Vessels-18 & Avg. \\
\hline
LLMDet & -- & \cellcolor{gray!10}54.4/70.8 & \cellcolor{gray!10}5.9/7.3 & \cellcolor{gray!10}10.7/22.4 & \cellcolor{gray!10}27.1/32.0 & \cellcolor{gray!10}9.5/13.5 & \cellcolor{gray!10}31.3/32.7 & \cellcolor{gray!10}3.5/5.1 & \cellcolor{gray!10}4.1/4.8 & \cellcolor{gray!10}18.3/23.6 \\
\hline

\multirow{3}{*}{
\begin{tabular}{@{}l@{}}
~~+~RSSD
\end{tabular}
}
& (i) OVD retention & -- & \cellcolor{red!10}39.0/54.4 & \cellcolor{red!10}38.2/50.5 & \cellcolor{red!10}45.2/60.3 & \cellcolor{red!10}43.9/58.8 & \cellcolor{red!10}45.3/59.8 & \cellcolor{red!10}31.3/42.9 & \cellcolor{red!10}30.4/40.4 & \cellcolor{red!10}39.0/52.4 \\
& (ii) Current-task acquisition & -- & \cellcolor{yellow!20}87.4/94.8 & \cellcolor{yellow!20}45.6/56.3 & \cellcolor{yellow!20}73.8/81.9 & \cellcolor{yellow!20}72.5/82.7 & \cellcolor{yellow!20}62.4/66.9 & \cellcolor{yellow!20}69.2/88.6 & \cellcolor{yellow!20}67.3/70.3 & \cellcolor{yellow!20}68.3/77.4 \\
& (iii) Final performance & \cellcolor{red!10}30.4/40.4 & 46.7/55.0 & 20.9/28.9 & 29.7/37.8 & 43.7/52.0 & 24.9/30.2 & 59.8/80.1 & \cellcolor{yellow!20}67.3/70.3 & \cellcolor{blue!15}40.4/49.3 \\
\hline

\multirow{3}{*}{
\begin{tabular}{@{}l@{}}
~~+~RSSD~+~KPD \\
(NoIn-Det)
\end{tabular}
}
& (i) OVD retention & -- & \cellcolor{red!10}49.8/65.1 & \cellcolor{red!10}49.5/64.3 & \cellcolor{red!10}49.1/63.7 & \cellcolor{red!10}47.0/62.0 & \cellcolor{red!10}47.1/61.9 & \cellcolor{red!10}45.7/60.2 & \cellcolor{red!10}41.7/54.8 & \cellcolor{red!10}47.1/61.7 \\
& (ii) Current-task acquisition & -- & \cellcolor{yellow!20}87.9/95.2 & \cellcolor{yellow!20}47.1/54.6 & \cellcolor{yellow!20}71.0/79.6 & \cellcolor{yellow!20}72.8/82.9 & \cellcolor{yellow!20}62.5/67.1 & \cellcolor{yellow!20}69.6/89.1 & \cellcolor{yellow!20}70.4/73.7 & \cellcolor{yellow!20}68.8/77.5 \\
& (iii) Final performance & \cellcolor{red!10}41.7/54.8 & 50.1/57.3 & 18.2/26.7 & 31.1/38.8 & 43.2/54.1 & 35.7/39.2 & 63.7/83.3 & \cellcolor{yellow!20}70.4/73.7 & \cellcolor{blue!15}44.3/53.5 \\
\hline
\end{tabular}}\vspace{-15pt}
\end{table*}

\begin{table*}[t!]
\centering
\tiny
\caption{Effect of the number of fine-tuned textual encoder layers in RSSD (\%).}
\label{tab:layer_ablation}
\resizebox{\textwidth}{!}{
\begin{tabular}{l|l|c|c|c|c|c|c|c|c|c}
\hline
Method & Protocol & MS COCO & Animals-25 & Clothing-6 & Cold Weapons-4 & Food-28 & Instruments-8 & Plants-25 & Vessels-18 & Avg. \\
\hline
LLMDet & -- & \cellcolor{gray!10}54.4/70.8 & \cellcolor{gray!10}5.9/7.3 & \cellcolor{gray!10}10.7/22.4 & \cellcolor{gray!10}27.1/32.0 & \cellcolor{gray!10}9.5/13.5 & \cellcolor{gray!10}31.3/32.7 & \cellcolor{gray!10}3.5/5.1 & \cellcolor{gray!10}4.1/4.8 & \cellcolor{gray!10}18.3/23.6 \\
\hline
\multirow{3}{*}{
\begin{tabular}{@{}l@{}}
NoIn-Det \\
(12 layers)
\end{tabular}}
& (i) OVD retention &--  & \cellcolor{red!10}41.2/53.0 & \cellcolor{red!10}38.1/53.9 & \cellcolor{red!10}37.6/49.0 & \cellcolor{red!10}39.3/51.5 & \cellcolor{red!10}40.7/52.8 & \cellcolor{red!10}40.8/54.1 & \cellcolor{red!10}20.3/28.3 &\cellcolor{red!10} 36.9 / 48.9 \\
& (ii) Current-task acquisition & -- & \cellcolor{yellow!20}86.9/94.6 & \cellcolor{yellow!20}55.6/60.1 & \cellcolor{yellow!20}74.8/82.4 & \cellcolor{yellow!20}73.4/83.7 & \cellcolor{yellow!20}59.9/66.0 & \cellcolor{yellow!20}69.9/89.1 & \cellcolor{yellow!20}68.3/70.2 &  \cellcolor{yellow!20}69.8/78.0 \\
& (iii) Final performance & \cellcolor{red!10}20.3/28.3 & 44.4/61.6 & 2.3/4.8 & 15.7/21.4 & 30.7/47.9 & 29.3/43.4 & 46.3/64.3 &\cellcolor{yellow!20} 68.3/70.2 & \cellcolor{blue!15}32.2/42.7 \\
\hline
\multirow{3}{*}{
\begin{tabular}{@{}l@{}}
NoIn-Det \\
(10 layers)
\end{tabular}}
& (i) OVD retention &--  & \cellcolor{red!10}43.9/58.1 & \cellcolor{red!10}39.3/51.6 & \cellcolor{red!10}43.5/56.6 & \cellcolor{red!10}41.5/55.0 & \cellcolor{red!10}43.4/56.4 & \cellcolor{red!10}43.0/57.8 & \cellcolor{red!10}34.2/44.8  &\cellcolor{red!10} 41.3/54.3 \\
& (ii) Current-task acquisition & -- & \cellcolor{yellow!20}88.8/96.0 & \cellcolor{yellow!20}55.4/63.0 & \cellcolor{yellow!20}71.1/79.8 & \cellcolor{yellow!20}72.2/82.4 & \cellcolor{yellow!20}66.2/70.4 & \cellcolor{yellow!20}70.4/89.6 & \cellcolor{yellow!20}67.9/70.9 & \cellcolor{yellow!20} 70.3/78.9 \\
& (iii) Final performance & \cellcolor{red!10}34.2/44.8 & 71.1/79.9 & 7.9/13.0 & 19.4/29.4 & 50.7/64.0 & 27.9/38.1 & 63.1/82.0 & \cellcolor{yellow!20}67.9/70.9 & \cellcolor{blue!15}42.8/52.8 \\
\hline
\multirow{3}{*}{
\begin{tabular}{@{}l@{}}
NoIn-Det \\
(8 layers)
\end{tabular}}
& (i) OVD retention &--  & \cellcolor{red!10}42.5/57.5 & \cellcolor{red!10}36.4/48.3 & \cellcolor{red!10}38.7/51.6 & \cellcolor{red!10}25.5/34.1 & \cellcolor{red!10}31.2/41.2 & \cellcolor{red!10}40.1/53.5 & \cellcolor{red!10}29.8/39.4  &\cellcolor{red!10} 34.9/46.5 \\
& (ii) Current-task acquisition & -- & \cellcolor{yellow!20}88.1/95.4 & \cellcolor{yellow!20}51.0/58.3 & \cellcolor{yellow!20}66.8/76.1 & \cellcolor{yellow!20}71.1/81.7 & \cellcolor{yellow!20}61.3/66.6 & \cellcolor{yellow!20}68.7/88.6 & \cellcolor{yellow!20}69.7/71.6 & \cellcolor{yellow!20} 68.1/76.9 \\
& (iii) Final performance & \cellcolor{red!10}29.8/39.4 & 61.7/72.6 & 14.3/19.7 & 27.7/39.5 & 48.4/58.1 & 32.4/39.5 & 56.5/74.9 & \cellcolor{yellow!20}69.7/71.6 & \cellcolor{blue!15}42.6/51.9 \\
\hline
\multirow{3}{*}{
\begin{tabular}{@{}l@{}}
NoIn-Det \\
(6 layers)
\end{tabular}}
& (i) OVD retention &--  & \cellcolor{red!10}48.4/63.7 & \cellcolor{red!10}42.8/56.1 & \cellcolor{red!10}50.3/65.7 & \cellcolor{red!10}40.7/55.2 & \cellcolor{red!10}49.8/65.2 & \cellcolor{red!10}40.5/55.4 & \cellcolor{red!10}41.4/55.3  &\cellcolor{red!10} 44.8/59.5 \\
& (ii) Current-task acquisition & -- & \cellcolor{yellow!20}87.4/94.4 & \cellcolor{yellow!20}52.4/55.9 & \cellcolor{yellow!20}72.3/80.3 & \cellcolor{yellow!20}72.6/82.6 & \cellcolor{yellow!20}62.8/66.4 & \cellcolor{yellow!20}69.8/89.3 & \cellcolor{yellow!20}66.3/70.1 &  \cellcolor{yellow!20}69.1/77.0 \\
& (iii) Final performance & \cellcolor{red!10}41.4/55.3 & 58.1/66.4 & 8.1/13.2 & 14.5/23.8 & 49.4/55.9 & 46.9/51.5 & 62.0/81.2 & \cellcolor{yellow!20}66.3/70.1 & \cellcolor{blue!15}43.3/52.2 \\
\hline
\multirow{3}{*}{
\begin{tabular}{@{}l@{}}
NoIn-Det \\
(4 layers)
\end{tabular}}
& (i) OVD retention &--  & \cellcolor{red!10}49.8/65.1 & \cellcolor{red!10}49.5/64.3 & \cellcolor{red!10}49.1/63.7 & \cellcolor{red!10}47.0/62.0 & \cellcolor{red!10}47.1/61.9 & \cellcolor{red!10}45.7/60.2 & \cellcolor{red!10}41.7/54.8  & \cellcolor{red!10}47.1/61.7 \\
& (ii) Current-task acquisition & -- & \cellcolor{yellow!20}87.9/95.2 & \cellcolor{yellow!20}47.1/54.6 & \cellcolor{yellow!20}71.0/79.6 & \cellcolor{yellow!20}72.8/82.9 & \cellcolor{yellow!20}62.5/67.1 & \cellcolor{yellow!20}69.6/89.1 & \cellcolor{yellow!20}70.4/73.7 &  \cellcolor{yellow!20}68.8/77.5 \\
& (iii) Final performance & \cellcolor{red!10}41.7/54.8 & 50.1/57.3 & 18.2/26.7 & 31.1/38.8 & 43.2/54.1 & 35.7/39.2 & 63.7/83.3 &\cellcolor{yellow!20} 70.4/73.7 & \cellcolor{blue!15}44.3/53.5 \\
\hline
\multirow{3}{*}{
\begin{tabular}{@{}l@{}}
NoIn-Det \\
(2 layers)
\end{tabular}}
& (i) OVD retention &--  & \cellcolor{red!10}52.7/68.5 & \cellcolor{red!10}51.6/67.2 & \cellcolor{red!10}51.6/67.2 & \cellcolor{red!10}47.0/63.0 & \cellcolor{red!10}50.0/65.5 & \cellcolor{red!10}46.3/60.8 & \cellcolor{red!10}40.4/53.3  & \cellcolor{red!10}48.5/63.6 \\
& (ii) Current-task acquisition & -- & \cellcolor{yellow!20}86.6/94.2 & \cellcolor{yellow!20}47.9/51.9 & \cellcolor{yellow!20}71.4/80.2 & \cellcolor{yellow!20}68.6/78.8 & \cellcolor{yellow!20}54.9/59.0 & \cellcolor{yellow!20}69.0/88.4 & \cellcolor{yellow!20}69.0/71.7 &  \cellcolor{yellow!20}66.8/74.9 \\
& (iii) Final performance & \cellcolor{red!10}40.4/53.3 & 19.4/27.0 & 5.7/8.9 & 15.9/20.9 & 40.4/48.6 & 31.6/37.2 & 63.0/81.9 &\cellcolor{yellow!20} 69.0/71.7 & \cellcolor{blue!15}35.7/43.7 \\
\hline
\multirow{3}{*}{
\begin{tabular}{@{}l@{}}
NoIn-Det \\
(0 layers)
\end{tabular}}
& (i) OVD retention &--  & \cellcolor{red!10}53.7/70.0 & \cellcolor{red!10}53.5/69.7 & \cellcolor{red!10}53.5/69.7 & \cellcolor{red!10}52.5/68.2 & \cellcolor{red!10}52.9/68.9 & \cellcolor{red!10}52.3/68.1 & \cellcolor{red!10}50.2/65.7  &\cellcolor{red!10} 52.7/68.6 \\
& (ii) Current-task acquisition & -- & \cellcolor{yellow!20}27.0/31.0 & \cellcolor{yellow!20}10.1/21.8 & \cellcolor{yellow!20}24.6/30.1 & \cellcolor{yellow!20}19.3/24.4 & \cellcolor{yellow!20}20.4/22.9 & \cellcolor{yellow!20}5.3/7.5 & \cellcolor{yellow!20}7.4/8.6 &  \cellcolor{yellow!20}16.3/20.9 \\
& (iii) Final performance & \cellcolor{red!10}50.2/65.7 & 4.7/8.2 & 1.5/3.8 & 6.4/9.8 & 18.7/24.2 & 9.0/13.8 & 3.6/5.5 & \cellcolor{yellow!20}7.4/8.6 & \cellcolor{blue!15}12.7/17.5 \\
\hline
\end{tabular}}\vspace{-15pt}
\end{table*}

\begin{table*}[t!]
\centering
\tiny
\caption{Sensitivity analysis of the update ratio $r$ in KPD on MS COCO and Novel-114 (\%).}
\label{tab:r_sensitivity}
\resizebox{\textwidth}{!}{
\begin{tabular}{l|l|c|c|c|c|c|c|c|c|c}
\hline
Method & Protocol & MS COCO & Animals-25 & Clothing-6 & Cold Weapons-4 & Food-28 & Instruments-8 & Plants-25 & Vessels-18 & Avg. \\
\hline
LLMDet & -- & \cellcolor{gray!10}54.4/70.8 & \cellcolor{gray!10}5.9/7.3 & \cellcolor{gray!10}10.7/22.4 & \cellcolor{gray!10}27.1/32.0 & \cellcolor{gray!10}9.5/13.5 & \cellcolor{gray!10}31.3/32.7 & \cellcolor{gray!10}3.5/5.1 & \cellcolor{gray!10}4.1/4.8 & \cellcolor{gray!10}18.3/23.6 \\
\hline

\multirow{3}{*}{
\begin{tabular}{@{}l@{}}
NoIn-Det \\
($r$=0.1)
\end{tabular}
}
& (i) OVD retention & -- & \cellcolor{red!10}41.5/55.5 & \cellcolor{red!10}45.4/59.2 & \cellcolor{red!10}48.4/63.1 & \cellcolor{red!10}39.8/53.8 & \cellcolor{red!10}44.4/58.6 & \cellcolor{red!10}31.2/43.4 & \cellcolor{red!10}41.2/54.5 & \cellcolor{red!10}41.7/55.4 \\
& (ii) Current-task acquisition & -- & \cellcolor{yellow!20}87.9/95.1 & \cellcolor{yellow!20}47.0/54.5 & \cellcolor{yellow!20}68.7/78.3 & \cellcolor{yellow!20}71.5/81.7 & \cellcolor{yellow!20}65.8/72.7 & \cellcolor{yellow!20}69.5/89.2 & \cellcolor{yellow!20}61.6/63.9 & \cellcolor{yellow!20}67.4/76.5 \\
& (iii) Final performance & \cellcolor{red!10}41.2/54.5 & 51.9/60.4 & 1.7/5.8 & 18.6/28.9 & 52.9/62.8 & 29.2/33.9 & 67.0/86.5 & \cellcolor{yellow!20}61.6/63.9 & \cellcolor{blue!15}36.2/44.6 \\
\hline

\multirow{3}{*}{
\begin{tabular}{@{}l@{}}
NoIn-Det \\
($r$=0.3)
\end{tabular}
}
& (i) OVD retention & -- & \cellcolor{red!10}38.6/52.6 & \cellcolor{red!10}35.7/47.0 & \cellcolor{red!10}43.2/56.7 & \cellcolor{red!10}40.7/55.0 & \cellcolor{red!10}41.4/55.4 & \cellcolor{red!10}37.6/51.0 & \cellcolor{red!10}39.0/52.5 & \cellcolor{red!10}39.5/52.9 \\
& (ii) Current-task acquisition & -- & \cellcolor{yellow!20}86.6/93.5 & \cellcolor{yellow!20}53.6/60.3 & \cellcolor{yellow!20}71.8/80.5 & \cellcolor{yellow!20}69.0/79.5 & \cellcolor{yellow!20}55.1/60.6 & \cellcolor{yellow!20}69.6/88.9 & \cellcolor{yellow!20}64.2/67.4 & \cellcolor{yellow!20}67.1/75.8 \\
& (iii) Final performance & \cellcolor{red!10}39.0/52.5 & 44.4/53.9 & 4.0/8.5 & 15.5/24.6 & 37.0/51.6 & 28.2/31.9 & 63.5/82.4 & \cellcolor{yellow!20}64.2/67.4 & \cellcolor{blue!15}37.0/46.6 \\
\hline

\multirow{3}{*}{
\begin{tabular}{@{}l@{}}
NoIn-Det \\
($r$=0.5)
\end{tabular}
}
& (i) OVD retention & -- & \cellcolor{red!10}49.8/65.1 & \cellcolor{red!10}49.5/64.3 & \cellcolor{red!10}49.1/63.7 & \cellcolor{red!10}47.0/62.0 & \cellcolor{red!10}47.1/61.9 & \cellcolor{red!10}45.7/60.2 & \cellcolor{red!10}41.7/54.8 & \cellcolor{red!10}47.1/61.7 \\
& (ii) Current-task acquisition & -- & \cellcolor{yellow!20}87.9/95.2 & \cellcolor{yellow!20}47.1/54.6 & \cellcolor{yellow!20}71.0/79.6 & \cellcolor{yellow!20}72.8/82.9 & \cellcolor{yellow!20}62.5/67.1 & \cellcolor{yellow!20}69.6/89.1 & \cellcolor{yellow!20}70.4/73.7 & \cellcolor{yellow!20}68.8/77.5 \\
& (iii) Final performance & \cellcolor{red!10}41.7/54.8 & 50.1/57.3 & 18.2/26.7 & 31.1/38.8 & 43.2/54.1 & 35.7/39.2 & 63.7/83.3 & \cellcolor{yellow!20}70.4/73.7 & \cellcolor{blue!15}44.3/53.5 \\
\hline

\multirow{3}{*}{
\begin{tabular}{@{}l@{}}
NoIn-Det \\
($r$=0.6)
\end{tabular}
}
& (i) OVD retention & -- & \cellcolor{red!10}48.2/63.8 & \cellcolor{red!10}46.7/61.1 & \cellcolor{red!10}47.3/62.0 & \cellcolor{red!10}47.9/63.7 & \cellcolor{red!10}48.5/64.2 & \cellcolor{red!10}41.4/55.7 & \cellcolor{red!10}40.2/53.0 & \cellcolor{red!10}45.7/60.5 \\
& (ii) Current-task acquisition & -- & \cellcolor{yellow!20}86.6/94.3 & \cellcolor{yellow!20}54.4/63.5 & \cellcolor{yellow!20}71.5/80.7 & \cellcolor{yellow!20}71.6/81.5 & \cellcolor{yellow!20}62.2/65.8 & \cellcolor{yellow!20}69.7/89.5 & \cellcolor{yellow!20}69.0/71.5 & \cellcolor{yellow!20}69.3/78.1 \\
& (iii) Final performance & \cellcolor{red!10}40.2/53.0 & 48.8/56.8 & 5.4/10.7 & 10.4/17.5 & 49.9/59.7 & 34.8/39.1 & 65.3/84.4 & \cellcolor{yellow!20}69.0/71.5 & \cellcolor{blue!15}40.5/49.1 \\
\hline

\multirow{3}{*}{
\begin{tabular}{@{}l@{}}
NoIn-Det \\
($r$=0.8)
\end{tabular}
}
& (i) OVD retention & -- & \cellcolor{red!10}47.5/62.1 & \cellcolor{red!10}41.1/54.1 & \cellcolor{red!10}46.7/60.7 & \cellcolor{red!10}48.9/64.0 & \cellcolor{red!10}48.7/63.6 & \cellcolor{red!10}42.0/55.5 & \cellcolor{red!10}39.9/52.9 & \cellcolor{red!10}45.0/59.0 \\
& (ii) Current-task acquisition & -- & \cellcolor{yellow!20}87.5/94.9 & \cellcolor{yellow!20}50.6/56.2 & \cellcolor{yellow!20}68.3/78.0 & \cellcolor{yellow!20}72.1/82.3 & \cellcolor{yellow!20}58.9/63.9 & \cellcolor{yellow!20}69.4/88.4 & \cellcolor{yellow!20}62.4/65.0 & \cellcolor{yellow!20}67.0/75.5 \\
& (iii) Final performance & \cellcolor{red!10}39.9/52.9 & 59.9/67.6 & 4.2/7.2 & 21.1/29.3 & 39.7/49.6 & 24.8/28.3 & 66.0/85.1 & \cellcolor{yellow!20}62.4/65.0 & \cellcolor{blue!15}39.7/48.1 \\
\hline

\multirow{3}{*}{
\begin{tabular}{@{}l@{}}
NoIn-Det \\
($r$=1.0)
\end{tabular}
}
& (i) OVD retention & -- & \cellcolor{red!10}39.0/54.4 & \cellcolor{red!10}38.2/50.5 & \cellcolor{red!10}45.2/60.3 & \cellcolor{red!10}43.9/58.8 & \cellcolor{red!10}45.3/59.8 & \cellcolor{red!10}31.3/42.9 & \cellcolor{red!10}30.4/40.4 & \cellcolor{red!10}39.0/52.4 \\
& (ii) Current-task acquisition & -- & \cellcolor{yellow!20}87.4/94.8 & \cellcolor{yellow!20}45.6/56.3 & \cellcolor{yellow!20}73.8/81.9 & \cellcolor{yellow!20}72.5/82.7 & \cellcolor{yellow!20}62.4/66.9 & \cellcolor{yellow!20}69.2/88.6 & \cellcolor{yellow!20}67.3/70.3 & \cellcolor{yellow!20}68.3/77.4 \\
& (iii) Final performance & \cellcolor{red!10}30.4/40.4 & 46.7/55.0 & 20.9/28.9 & 29.7/37.8 & 43.7/52.0 & 24.9/30.2 & 59.8/80.1 & \cellcolor{yellow!20}67.3/70.3 & \cellcolor{blue!15}40.4/49.3 \\
\hline
\end{tabular}}\vspace{-15pt}
\end{table*}

\section{Conclusion}
In this paper, we introduce COVD, a new task that continually injects novel concepts while preserving previously injected concepts and original OVD knowledge. To support COVD, we construct Novel-114, a benchmark with 114 novel concepts organized into seven sequential stages.
We propose NoIn-Det, a parameter-efficient framework that freezes the visual encoder and selectively updates the textual branch. With Representation Space Stability Distillation (RSSD) and Knowledge-aware Parameter Decoupling (KPD), NoIn-Det stabilizes the semantic space and constrains old-knowledge-sensitive parameters. Experiments on Novel-114 and COCO show that NoIn-Det effectively balances novel concept learning, historical concept retention, and original OVD knowledge preservation, consistently outperforming existing continual learning methods. 
{
\small
\bibliographystyle{unsrt}
\bibliography{main}
}

\newpage

\appendix
\section*{Appendix}

\section{Related Works}
\label{sec:Related}
\textbf{Open-vocabulary object detection (OVD)} \cite{zareian2021open} aims to detect novel object categories that are unseen during training. In recent years, with the rapid development of VLMs such as CLIP~\cite{radford2021learning} and ALIGN~\cite{jia2021scaling}, OVD has become an important research direction, enabling models to recognize both base and novel categories within a unified cross-modal semantic space.
Existing OVD methods can be roughly grouped into four categories. The first transfers open semantic knowledge from VLMs into detectors through knowledge distillation, such as ViLD~\cite{gu2021open}, DetPro~\cite{du2022learning}, and methods with enhanced semantic modeling like OADP~\cite{wang2023object} and DK-DETR~\cite{li2023distilling}. The second relies on pseudo labels or weakly supervised object mining, including Detic~\cite{zhou2022detecting}, OCO~\cite{bangalath2022bridging}, ProxyDet~\cite{jeong2024proxydet}, LBP~\cite{li2024learning}, SAS-Det~\cite{zhao2024taming}, and OV-DQUO~\cite{wang2025ov}, which mine potential objects using image-level labels, class-agnostic detectors, or proxy categories. The third builds detection frameworks on frozen VLMs and trains only the detection head or a small number of downstream modules, as in F-VLM~\cite{kuo2022f}, CLIPSelf~\cite{wu2023clipself}, DST-Det~\cite{xu2024dst}, and DeCLIP~\cite{wang2025declip}. The fourth learns stronger region-level vision-language alignment from large-scale region-text pair data, including RO-ViT~\cite{kim2023region}, CORA~\cite{wu2023cora}, YOLO-World~\cite{cheng2024yolo}, YOLOE~\cite{wang2025yoloe}, GLIP~\cite{li2022grounded}, GroundingDINO~\cite{liu2024grounding}, and LLMDet~\cite{fu2025llmdet}.
In this work, we build on LLMDet, which is pre-trained on large-scale image region-text pairs, and further study continual injection of novel concepts on top of this strong OVD foundation.

\textbf{Continual learning for VLMs} aims to incrementally acquire novel categories from sequential data streams while mitigating catastrophic forgetting. Beyond maintaining recognition of both old and new categories, it also requires preserving the cross-modal alignment and zero-shot transfer ability inherited from pre-training.
Existing methods mainly follow four directions. First, large-scale distillation methods, such as ZSCL~\cite{zheng2023preventing}, preserve prior knowledge through distillation on a large reference dataset and weight ensembling, but require substantial external data and high computation. Second, prompt-based methods, such as DIKI~\cite{tang2024mind}, adapt frozen backbones with residual prompts, yet the acquired knowledge is largely stored in additional prompts rather than stably internalized, leading to growing storage and management overhead as tasks accumulate. Third, parameterized-module methods absorb new knowledge through additional structures, including adapters in SGVF~\cite{cheng2026incremental} and LADA~\cite{luo2025lada}, and mixture-of-experts in MoE-Adapters~\cite{yu2024boosting}; although effective, these methods inevitably introduce continual parameter growth. Fourth, parameter-efficient tuning and re-parameterization methods, such as C-CLIP~\cite{luo2026keeplora} and ZiRa~\cite{deng2024zero}, improve adaptation efficiency but may still perturb the original semantic space during continual updating, weakening the stability of old knowledge representations.
Overall, existing approaches mainly rely on peripheral modules or local parameter adaptation, rather than stably internalizing novel concepts into the model itself. As a result, they still struggle to balance new knowledge acquisition, old knowledge preservation, and retention of pre-trained capabilities. Another key difference lies in the adaptation branch: SGVF~\cite{cheng2026incremental} and LADA~\cite{luo2025lada} mainly adapt the visual branch, while most other methods modify both visual and textual branches. In contrast, our method updates only the textual branch, motivated by the observation that the pre-trained visual encoder already provides strong and transferable visual representations, whereas continual tuning of the visual branch is more likely to damage its complex and generalizable capacity.

\section{More Details of Novel-114}
\label{sec:More_details_novel114}

\subsection{Task-wise Concept Composition}
\label{sec:Concepts}
Table~\ref{tab:114} presents the task-wise concept composition of Novel-114. The seven tasks cover diverse concept domains with clear scenario and semantic differences, including animals, clothing, cold weapons, food, instruments, plants, and vessels. These groups span natural objects, man-made objects, specialized concepts, and daily-life concepts, with task sizes ranging from small concept groups to larger ones. 
Importantly, the task order is not organized by category number, visual difficulty, or semantic similarity, but follows a heterogeneous cross-scenario injection sequence. This design better reflects realistic deployment scenarios, where novel concepts may arrive from different domains and in non-uniform batches. Therefore, Novel-114 not only supports continual concept injection, but also provides a challenging benchmark for evaluating cross-scenario generalization, new concept acquisition, and old knowledge retention.

\begin{table*}[t!]
\centering
\scriptsize
\caption{Task-wise category composition of Novel-114.}  
\label{tab:114}
\renewcommand{\arraystretch}{1.18}
\setlength{\tabcolsep}{5pt}
\begin{tabularx}{\textwidth}{l X}
\toprule
\textbf{Task} & \textbf{Concepts} \\
\midrule
Animals-25 &
Rupicola Peruvianus,
Alpheus,
Okapia Johnstoni,
Andrias Japonicus,
Dryococelus Australis,
Nipponia Nippon,
Megadyptes Antipodes,
Muntiacus Muntjak,
Nephrurus Amyae,
Dipodidae,
Pacu Fish,
Penelopides Panini,
Hyalinobatrachium Pellucidum,
Varecia Variegata,
Moloch Horridus Gray,
Uroplatus Phantasticus,
Chrysiridia Rhipheus,
Tapiridae,
Odontodactylus Scyllarus,
Litocranius,
Lipotes,
Opisthocomus Hoazin,
Leontopithecus Rosalia,
Cebidae,
Setonix Brachyurus \\
\midrule

Clothing-6 &
CoatSkirt,
RuQun,
ZhuZiShenYi,
QuJu,
ZhiDuo,
ZhiJu \\
\midrule

Cold Weapons-4 &
Mao,
Ge,
Yue,
Zu \\
\midrule

Food-28 &
Stew,
Dried Fish,
Haupia,
Three Cup Chicken,
Zoni,
Chip Butty,
Cold Tofu,
Brownie,
Mung Bean Cake,
Zha Jiang Mian,
Dry Curry,
Mozuku,
Namero,
Lightly Roasted Fish,
Ganmodoki,
Kushikatu,
Chop Suey,
Goya Chanpuru,
Kamameshi,
Oshiruko,
Zeng Cake,
Nanbanzuke,
Trunip Pudding,
Samul,
BianZhong,
Ruan,
KongHou,
HuLuSi \\
\midrule

Instruments-8 &
BianQing,
Nao,
Sheng,
Se,
Xun,
Tao,
Duo,
Bo \\
\midrule

Plants-25 &
Abutilon Pictum,
Conophytum Caculus,
Dionaea Muscipula,
Saussurea Involucrata,
Thysanocarpus Radians,
Ochna KirkiiOliv,
Conophytum Burgeri,
Diplocyclos Palmatus,
Ecballium Elaterium,
Peristeria Elata Hook,
Hedysarum,
Cookeina Speciosa,
Adenium Obesum,
Gentianella Hirculus,
Rafflesia Arnoldi,
Saussurea Gossipiphora,
Davidia Involucrata,
Darlingtonia Californica,
Centropogon Cornutus,
Nepenthes Distillatoria,
Sauropus Androgynus,
Euphorbia Obesa,
Dolichandrone Stipulata,
Rhodomyrtus Tomentosa,
Lasiosphaera \\
\midrule

Vessels-18 &
Xu,
Gui,
Li,
Ding,
Jue,
Jia,
Yi,
Zun,
Gu,
Gong,
You,
Bu,
Fu,
He,
Lei,
Zhi,
Yan,
Fou \\
\bottomrule
\end{tabularx}\vspace{-10pt}
\end{table*}

\begin{figure*}[t!]
	\centering
	\includegraphics[width=0.96\linewidth]{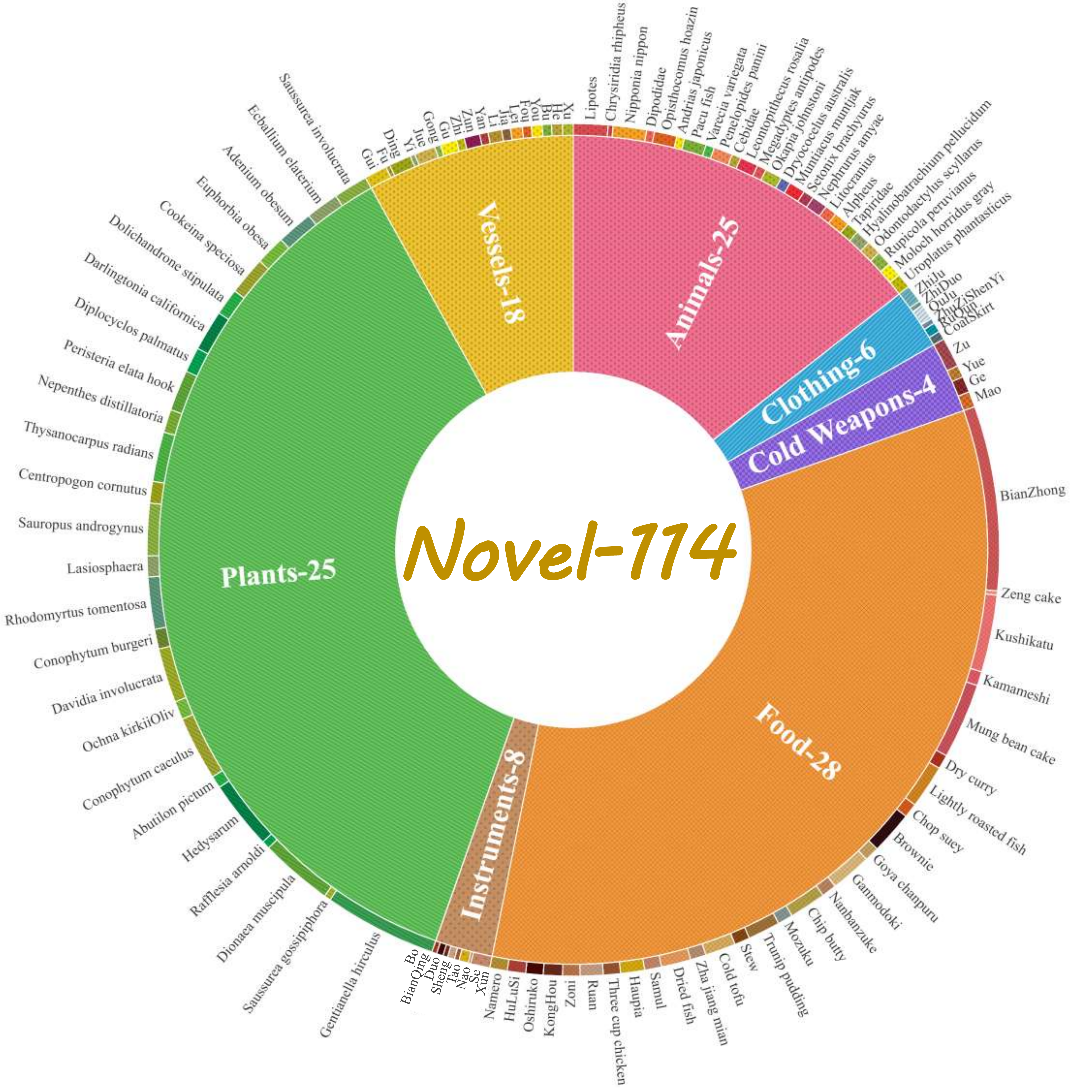}  \vspace{-10pt}
	\caption{Sunburst visualization of the task-wise composition and data distribution of Novel-114. The inner ring denotes the seven continual learning tasks, the outer ring shows their concepts, and sector areas reflect relative data scales.}  
	\label{fig:sunburst}
	\vspace{-10pt}
\end{figure*}

\begin{figure*}[t!]
	\centering
	\includegraphics[width=1.0\linewidth]{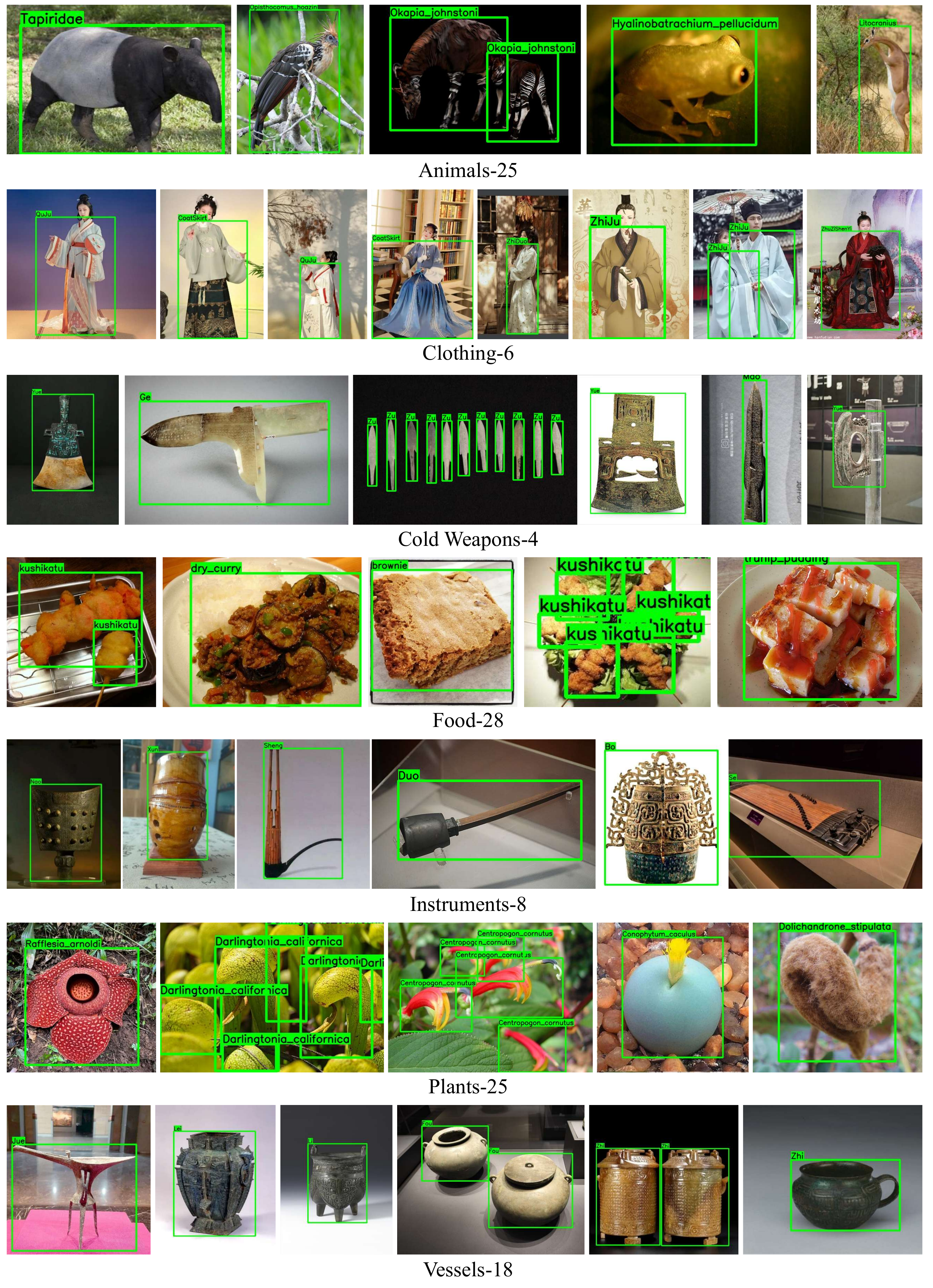}\vspace{-10pt}
	\caption{Visualization of Novel-114. The benchmark contains diverse novel, rare, and specialized concepts that are difficult to cover with existing open-vocabulary knowledge, while complex backgrounds and scale variations further increase the detection challenge.}
	\label{fig:novel-114}
	\vspace{-10pt}
\end{figure*}

\subsection{Sunburst Visualization}
\label{sec:Sunburst}
Fig.~\ref{fig:sunburst} provides an intuitive visualization of the hierarchical composition and data distribution of Novel-114. The inner ring denotes the seven continual learning tasks, and the outer ring shows their corresponding concepts. The sector area reflects the relative data scale of each task/concept, showing that Novel-114 contains both small and large concept groups. Together with the distinct semantic domains, this non-uniform structure better simulates realistic continual learning scenarios, where novel concepts arrive from different domains and with varying data scales.

\subsection{Visualization Examples}
\label{sec:Examples}
We visualize Novel-114 in Fig.~\ref{fig:novel-114}. The benchmark covers diverse novel concepts across animals, clothing, cold weapons, food, instruments, plants, and vessels, spanning natural objects, man-made objects, specialized concepts, and daily-life concepts.

As shown in the figure, Novel-114 exhibits rich variation in object appearance, scale, pose, and context. Some categories require fine-grained discrimination among visually similar concepts, while others appear in cluttered scenes, unusual viewpoints, or challenging lighting conditions. Small, sparse, and partially occluded instances further increase the difficulty of localization and recognition.
These properties make Novel-114 a challenging benchmark for continual novel concept injection: models must acquire genuinely novel concepts, distinguish them from related known semantics, and remain robust across diverse visual scenarios.

\section{More Results}
\label{sec:More_results}
\begin{wraptable}{r}{0.57\textwidth} \vspace{-19pt}
\centering
\tiny
\caption{Robustness to common concept sampling (\%).}
\setlength{\tabcolsep}{1.6pt}
\label{tab:Common_robust}
\begin{tabular}{l|l|c|c|c|c|c}
\hline
Method & Protocol  & Set 1 & Set 2 & Set 3 & Set 4 & Set 5  \\
\hline
\multirow{3}{*}{NoIn-Det}
&(i) OVD retention &\cellcolor{red!10}47.3/60.9&\cellcolor{red!10}46.8/60.6&\cellcolor{red!10}47.1/62.0&\cellcolor{red!10}46.6/60.9&\cellcolor{red!10}47.0/61.1 \\
&(ii) Current-task acquisition &\cellcolor{yellow!20}68.6/77.1&\cellcolor{yellow!20}68.7/78.1&\cellcolor{yellow!20}68.3/77.6&\cellcolor{yellow!20}67.9/76.9&\cellcolor{yellow!20}69.1/77.8 \\
&(iii) Final performance &\cellcolor{blue!15}44.1/53.0&\cellcolor{blue!15}43.8/52.8&\cellcolor{blue!15}43.9/53.2&\cellcolor{blue!15}44.5/53.2&\cellcolor{blue!15}44.1/52.7 \\
\hline
\end{tabular}\vspace{-15pt}
\end{wraptable}
\subsection{Robustness to Common Concept Sampling}
\label{sec:Common_Concept_Sampling}
Since RSSD uses randomly sampled common concepts as lightweight textual old knowledge, we further evaluate whether the performance is sensitive to the sampled common set. We randomly construct five Common-100 sets from ImageNet~\cite{deng2009imagenet} and Objects365~\cite{shao2019objects365}, while keeping all other settings unchanged. As shown in Table~\ref{tab:common_robust}, NoIn-Det achieves stable performance across different common sets, with only minor variance in both COCO retention and Novel-114 performance. This indicates that RSSD does not rely on a specific common concept set, and the sampled common texts mainly serve as general semantic anchors for stabilizing the textual representation space.

\subsection{Second-round Concept Injection}
\label{sec:Second}
To further evaluate the scalability of our method, we conduct a second round of concept injection by initializing from the final model obtained after the first round and repeating the same training protocol on Novel-114.
As shown in Table~\ref{tab:second_round}, our method continues to achieve strong performance under extended continual learning. In particular, it obtains the best final average performance of 55.1\%/64.2\%, clearly outperforming SGVF (49.2\%/56.3\%), ZiRa (44.8\%/52.0\%), and MoE-Adapters (39.5\%/45.3\%).

\textbf{Knowledge accumulation.}
Compared with the first round, our method improves from 44.3\%/53.5\% to 55.1\%/64.2\% in final average performance, indicating that the model can accumulate knowledge across multiple injection rounds rather than merely relearning or overwriting previous concepts.

\textbf{Retention and stability.}
Despite undergoing an additional round of concept injection, our method maintains stronger original OVD retention than competing methods. After the final second-round task, our method still achieves 39.5\%/52.5\% on MS COCO, while SGVF and ZiRa drop to 16.9\%/23.3\% and 15.0\%/19.5\%, respectively. This further confirms that RSSD and KPD effectively mitigate catastrophic forgetting in longer continual learning processes.

\textbf{Comparison with existing methods.}
Most baseline methods show limited robustness under repeated knowledge injection. For example, KeepLoRA obtains only 16.1\%/21.9\% final average performance, while C-CLIP reaches 27.6\%/34.0\%. In contrast, our method maintains both strong plasticity and stability, leading to consistent performance improvements.

\textbf{Overall analysis.}
These results demonstrate that our method is not only effective for single-round continual learning, but also scalable to multi-round concept injection. This highlights its potential for real-world scenarios where new concepts are continuously introduced over time.


\begin{table*}[t!]
\centering
\tiny
\caption{Second-round concept injection on MS COCO and Novel-114 (\%).}
\vspace{1pt}
\label{tab:second_round}
\resizebox{\textwidth}{!}{
\begin{tabular}{l|l|c|c|c|c|c|c|c|c|c}
\hline
\multirow{2}{*}{Method} & \multirow{2}{*}{\diagbox[width=8em]{Train}{Test}} & \multirow{2}{*}{MS COCO} & \multirow{2}{*}{Animals-25} & \multirow{2}{*}{Clothing-6} & \multirow{2}{*}{Cold Weapons-4} & \multirow{2}{*}{Food-28} & \multirow{2}{*}{Instruments-8} & \multirow{2}{*}{Plants-25} & \multirow{2}{*}{Vessels-18} & \multirow{2}{*}{Avg.} \\
 &  & &  &  &  &  &  &  &  & \\
\hline
\rowcolor{gray!15} LLMDet &--  & 54.4/70.8 & 5.9/7.3   & 10.7/22.4 & 27.1/32.0 & 9.5/13.5  & 31.3/32.7 & 3.5/5.1   & 4.1/4.8   & 18.3/23.6 \\
\hline
\multirow{8}{*}{ZiRa}
& Animals-25      &\cellcolor{red!10}18.9/24.7 &\cellcolor{yellow!20}87.7/95.3&16.3/24.8&22.6/27.1&2.6/4.0&18.4/21.7&47.6/62.0&37.2/40.8&\cellcolor{blue!15}31.4/37.6 \\
& Clothing-6      &\cellcolor{red!10}17.2/22.5 &85.3/93.2&\cellcolor{yellow!20}46.4/50.3&20.4/26.1&43.3/50.7&15.2/19.6&46.3/60.0&41.2/44.6&\cellcolor{blue!15}39.4/45.9\\
& Cold Weapons-4  &\cellcolor{red!10}18.8/24.5 &86.1/94.2&46.0/51.6&\cellcolor{yellow!20}66.0/73.6&42.3/49.4&12.3/15.7&45.5/58.9&30.2/35.0&\cellcolor{blue!15}43.4/50.4 \\
& Food-28         &\cellcolor{red!10}20.4/26.5 &83.7/91.7&27.5/30.3&43.5/51.9&\cellcolor{yellow!20}73.2/83.7&24.1/28.6&43.9/57.0&32.8/35.3&\cellcolor{blue!15}43.6/50.6 \\
& Instruments-8   &\cellcolor{red!10}21.7/28.0 &84.2/92.2&21.5/24.7&43.6/51.1&71.2/81.4&\cellcolor{yellow!20}48.1/53.2&45.3/59.1&33.1/35.4&\cellcolor{blue!15}46.1/53.1 \\
& Plants-25       &\cellcolor{red!10}17.3/22.6 &65.8/72.5&13.7/18.7&43.5/51.0&64.7/74.3&44.9/50.5&\cellcolor{yellow!20}68.7/87.9&33.8/36.1&\cellcolor{blue!15}44.1/51.7 \\
& Vessels-18      &\cellcolor{red!10}15.0/19.5 &65.5/72.1&13.5/16.6&43.6/51.0&57.2/67.4&36.3/40.0&65.1/83.7&\cellcolor{yellow!20}62.1/65.3&\cellcolor{blue!15}44.8/52.0 \\
\hline
\multirow{8}{*}{MoE-Adapters}
& Animals-25      &\cellcolor{red!10} 33.4/44.9 &\cellcolor{yellow!20} 85.5/92.7 & 12.8/23.2 & 22.7/26.5 & 8.0/10.0  & 11.3/12.0 & 2.3/3.7   & 6.5/7.8   &\cellcolor{blue!15}22.8/27.6 \\
& Clothing-6      &\cellcolor{red!10} 30.3/41.0 & 84.5/92.1 &\cellcolor{yellow!20} 34.2/41.2 & 23.1/27.0 & 6.2/8.3   & 8.3/9.6   & 1.8/3.0   & 7.5/9.2   &\cellcolor{blue!15}24.5/28.9\\
& Cold Weapons-4  &\cellcolor{red!10} 28.4/38.4 & 82.0/89.9 & 33.3/43.5 &\cellcolor{yellow!20} 43.6/49.6 & 5.2/6.9   & 6.3/7.8   & 2.1/3.3   & 8.4/9.1   &\cellcolor{blue!15}26.2/31.1 \\
& Food-28         &\cellcolor{red!10} 31.2/42.8 & 78.9/86.9 & 32.7/40.4 & 46.5/53.7 &\cellcolor{yellow!20} 62.4/71.8 & 14.2/17.0 & 3.2/4.6   & 6.7/7.3   &\cellcolor{blue!15}34.5/40.6\\
& Instruments-8   &\cellcolor{red!10} 32.6/44.8 & 80.1/88.0 & 33.4/40.5 & 42.0/49.2 & 60.5/69.8 &\cellcolor{yellow!20} 23.4/25.5 & 3.8/5.2   & 6.4/7.1   &\cellcolor{blue!15}35.3/41.3\\
& Plants-25       &\cellcolor{red!10} 29.0/40.1 & 77.9/85.7 & 27.0/33.9 & 41.9/48.4 & 50.8/59.8 & 20.1/22.6 & \cellcolor{yellow!20}35.6/44.9 & 3.4/4.2   &\cellcolor{blue!15}35.7/42.5 \\
& Vessels-18      &\cellcolor{red!10} 28.2/38.8 & 66.1/73.0 & 25.6/28.6 & 33.5/40.0 & 48.0/56.1 & 22.1/23.7 & 34.1/43.3 & \cellcolor{yellow!20}58.1/59.1 &\cellcolor{blue!15}39.5/45.3\\
\hline
\multirow{8}{*}{LADA}
& Animals-25      & \cellcolor{red!10}31.9/43.2 & \cellcolor{yellow!20}87.3/95.6 & 13.0/23.0 & 28.8/32.5 & 27.7/33.9 & 22.8/24.9 & 24.3/32.1 & 21.7/23.1 &\cellcolor{blue!15}32.2/38.5\\
& Clothing-6      & \cellcolor{red!10}23.2/30.7 & 78.3/89.0 & \cellcolor{yellow!20}64.0/67.9 & 18.7/22.2 & 16.9/24.3 & 24.1/26.8 & 17.3/23.1 & 26.7/28.1 &\cellcolor{blue!15}33.7/39.0 \\
& Cold Weapons-4  & \cellcolor{red!10}22.5/30.2 & 72.7/81.8 & 38.7/44.7 & \cellcolor{yellow!20}75.1/82.6 & 15.9/20.6 & 11.2/15.1 & 16.2/21.8 & 18.0/19.0 &\cellcolor{blue!15}33.8/39.5 \\
& Food-28         & \cellcolor{red!10}27.6/37.4 & 61.3/70.6 & 19.2/24.8 & 41.0/47.8 & \cellcolor{yellow!20}74.5/85.3 & 28.2/31.3 & 6.9/9.9   & 6.0/6.7   &\cellcolor{blue!15}33.1/39.2 \\
& Instruments-8   & \cellcolor{red!10}18.0/24.3 & 43.4/51.4 & 25.8/31.7 & 40.5/46.9 & 58.3/69.2 &\cellcolor{yellow!20} 70.5/75.1 & 4.9/6.7   & 11.8/13.1 &\cellcolor{blue!15}34.2/39.8 \\
& Plants-25       & \cellcolor{red!10}20.8/28.7 & 43.3/52.3 & 19.3/25.9 & 35.6/43.7 & 45.4/54.7 & 33.9/38.5 & \cellcolor{yellow!20}70.4/89.1 & 10.4/12.0 &\cellcolor{blue!15}34.9/43.1 \\
& Vessels-18      & \cellcolor{red!10}17.6/23.9 & 35.8/41.0 & 14.1/17.1 & 26.7/28.6 & 20.2/26.0 & 21.9/24.2 & 49.6/66.3 &\cellcolor{yellow!20} 75.6/76.4 &\cellcolor{blue!15}32.7/37.9 \\
\hline
\multirow{8}{*}{C-CLIP}
& Animals-25       &\cellcolor{red!10}42.7/57.2&\cellcolor{yellow!20}75.2/82.6&25.8/39.0&30.6/36.2&21.8/27.1&12.4/14.2&10.2/12.9&7.8/8.7&\cellcolor{blue!15}28.3/34.7\\
& Clothing-6       &\cellcolor{red!10}41.9/55.9&56.5/63.1&\cellcolor{yellow!20}43.7/47.2&56.5/63.1&17.5/21.8&8.1/10.7&8.2/10.8&8.9/9.8&\cellcolor{blue!15}30.2/35.3 \\
& Cold Weapons-4   &\cellcolor{red!10}45.5/61.6&48.4/54.3&36.5/38.5&\cellcolor{yellow!20}50.3/59.1&21.3/26.6&20.3/24.0&7.6/10.0&12.7/14.5&\cellcolor{blue!15}30.3/36.1 \\
& Food-28          &\cellcolor{red!10}45.6/61.0&28.0/32.2&19.6/25.8&29.5/34.4&\cellcolor{yellow!20}56.0/65.4&31.1/33.3&6.7/8.9&6.4/7.4&\cellcolor{blue!15}27.9/33.6 \\
& Instruments-8    &\cellcolor{red!10}46.2/62.0&21.7/25.3&21.4/29.3&27.8/33.5&42.2/49.9&\cellcolor{yellow!20}44.1/46.9&6.8/8.9&6.9/9.1&\cellcolor{blue!15}27.1/33.1 \\
& Plants-25        &\cellcolor{red!10}38.3/52.7&19.1/22.4&22.1/29.0&34.7/43.0&33.3/39.9&40.7/53.0&\cellcolor{yellow!20}39.5/52.1&7.0/10.1&\cellcolor{blue!15}29.3/37.8 \\
& Vessels-18       &\cellcolor{red!10}32.1/44.9&19.9/23.2&25.6/33.1&43.6/50.0&18.7/23.4&16.1/20.1&33.0/44.2&\cellcolor{yellow!20}31.6/32.9&\cellcolor{blue!15}27.6/34.0 \\
\hline
\multirow{8}{*}{SGVF}
& Animals-25      &\cellcolor{red!10} 26.1/34.8 & \cellcolor{yellow!20}85.2/93.8 & 17.6/21.7 & 39.5/45.1 & 45.9/53.7 & 41.3/44.8 & 29.0/39.4 & 42.1/43.1 &\cellcolor{blue!15}40.8/47.1  \\
& Clothing-6      &\cellcolor{red!10} 24.6/32.9 & 83.3/92.3 & \cellcolor{yellow!20}67.3/72.5 & 38.3/44.2 & 41.0/48.5 & 46.3/50.5 & 28.6/39.9 & 32.8/34.0 &\cellcolor{blue!15}45.3/51.9 \\
& Cold Weapons-4  &\cellcolor{red!10} 23.6/31.7 & 81.9/90.4 & 48.6/53.4 & \cellcolor{yellow!20}74.1/81.4 & 39.6/46.1 & 31.8/35.1 & 25.5/35.4 & 23.2/25.8 &\cellcolor{blue!15}43.5/49.9 \\
& Food-28         &\cellcolor{red!10} 25.3/34.3 & 76.9/87.7 & 35.5/40.1 & 60.2/67.2 & \cellcolor{yellow!20}73.6/84.5 & 28.0/32.4 & 12.3/17.9 & 17.1/18.1 &\cellcolor{blue!15}41.1/47.8 \\
& Instruments-8   &\cellcolor{red!10} 21.8/29.8 & 71.2/82.6 & 34.2/37.6 & 50.1/57.3 & 71.4/81.9 &\cellcolor{yellow!20} 63.6/69.3 & 12.2/17.6 & 15.1/16.3 &\cellcolor{blue!15}42.5/49.1 \\
& Plants-25       &\cellcolor{red!10} 22.5/30.7 & 72.5/83.3 & 35.1/39.1 & 48.8/54.6 & 64.2/75.2 & 54.1/58.5 & \cellcolor{yellow!20}68.4/88.3 & 15.8/16.7 &\cellcolor{blue!15}47.7/55.8 \\
& Vessels-18      &\cellcolor{red!10} 16.9/23.3 & 69.3/78.8 & 27.8/29.7 & 46.1/51.3 & 60.0/70.2 & 43.6/48.6 & 50.5/68.4 & \cellcolor{yellow!20}79.4/80.3 &\cellcolor{blue!15}49.2/56.3 \\
\hline
\multirow{8}{*}{KeepLoRa}
& Animals-25      &\cellcolor{red!10} 43.6/58.9 & \cellcolor{yellow!20}7.7/11.2  & 15.8/32.0 & 28.5/34.5 & 10.2/15.5 & 27.9/31.3 & 3.3/5.0   & 6.0/7.5  & \cellcolor{blue!15}17.9/24.5 \\
& Clothing-6      &\cellcolor{red!10} 40.6/56.9 & 6.1/7.4   & \cellcolor{yellow!20}14.2/24.6 & 27.1/34.3 & 9.3/14.0  & 24.6/29.5 & 3.2/4.7   & 4.4/5.2  &\cellcolor{blue!15}16.2/22.1 \\
& Cold Weapons-4  &\cellcolor{red!10} 38.5/54.1 & 5.8/7.1   & 10.2/20.6 & \cellcolor{yellow!20}25.9/32.4 & 8.7/12.8  & 26.5/32.2 & 3.0/4.3   & 3.9/5.0  &\cellcolor{blue!15}15.3/21.1 \\
& Food-28         &\cellcolor{red!10} 46.0/62.0 & 7.3/9.1   & 9.4/21.2  & 29.1/36.4 & \cellcolor{yellow!20}16.0/22.2 & 27.7/32.1 & 3.6/5.3   & 4.5/5.6  &\cellcolor{blue!15}18.0/24.2 \\
& Instruments-8   &\cellcolor{red!10}40.0/56.0 & 6.0/7.5   & 8.4/18.2  & 23.0/28.7 & 8.6/13.2  & \cellcolor{yellow!20}26.0/28.6 & 3.1/4.6   & 4.2/5.3  &\cellcolor{blue!15}14.9/20.3 \\
& Plants-25       &\cellcolor{red!10}32.2/45.0 & 7.2/9.5   & 11.2/24.7 & 24.8/29.8 & 8.3/12.9  & 25.3/29.2 & \cellcolor{yellow!20}5.3/7.3   & 4.8/6.2  &\cellcolor{blue!15}14.9/20.6 \\
& Vessels-18      &\cellcolor{red!10}40.8/55.9 & 6.1/7.6   & 11.9/22.8 & 22.5/28.6 & 8.2/13.3  & 28.0/32.2 & 3.6/5.2   &\cellcolor{yellow!20} 7.9/9.7  &\cellcolor{blue!15}16.1/21.9  \\
\hline
\multirow{7}{*}{\begin{tabular}[l]{@{}l@{}}
NoIn-Det \\
\end{tabular}}
& Animals-25     &\cellcolor{red!10}46.6/62.0  &\cellcolor{yellow!20}88.4/95.9 &22.4/29.4  & 36.5/44.8  & 41.8/49.0  & 32.3/35.7  & 20.5/28.2   & 35.1/37.4   &\cellcolor{blue!15}40.5/47.8        \\
& Clothing-6     &\cellcolor{red!10}42.5/56.3   &75.9/90.8   &\cellcolor{yellow!20}59.8/65.4  &31.6/36.6   &35.9/46.4  &30.1/34.1  &17.1/25.9    & 33.5/35.8   &\cellcolor{blue!15}40.8/48.9        \\
& Cold Weapons-4 &\cellcolor{red!10}45.2/59.7   &76.9/91.5   &55.6/63.7  &\cellcolor{yellow!20}74.4/82.7  &44.3/54.5  &31.6/34.5  &19.2/28.0  &37.0/39.3        & \cellcolor{blue!15}48.0/56.7        \\
& Food-28        &\cellcolor{red!10}44.8/58.7   &81.7/90.4  &25.3/30.7  &57.2/64.1  &\cellcolor{yellow!20}72.6/83.0  &32.8/37.5 &28.9/38.8  &38.3/39.6   &\cellcolor{blue!15}47.7/55.4      \\
& Instruments-8  &\cellcolor{red!10}44.9/59.1   &74.8/88.4   &28.0/34.0  &62.9/69.4  &70.0/81.4  &\cellcolor{yellow!20}58.5/64.7 & 26.3/36.8  &38.1/39.9        & \cellcolor{blue!15}50.4/59.2        \\
& Plants-25      &\cellcolor{red!10}38.4/52.2   &80.9/89.2   &20.4/27.9  &51.9/58.8  &66.5/77.0  &64.6/73.1  &\cellcolor{yellow!20}71.2/90.1 & 41.0/43.4        & \cellcolor{blue!15}54.4/64.0        \\
& Vessels-18     &\cellcolor{red!10}39.5/52.5   &77.0/88.0   &27.9/34.5  &50.8/57.1  &61.5/73.6  &47.4/53.6  &66.2/86.0  &\cellcolor{yellow!20}70.1/72.1 & \cellcolor{blue!15}55.1/64.2  \\
\hline
\end{tabular} 
}\vspace{-0pt}
\end{table*}

\begin{figure*}[t!]
	\centering
	\includegraphics[width=1.0\linewidth]{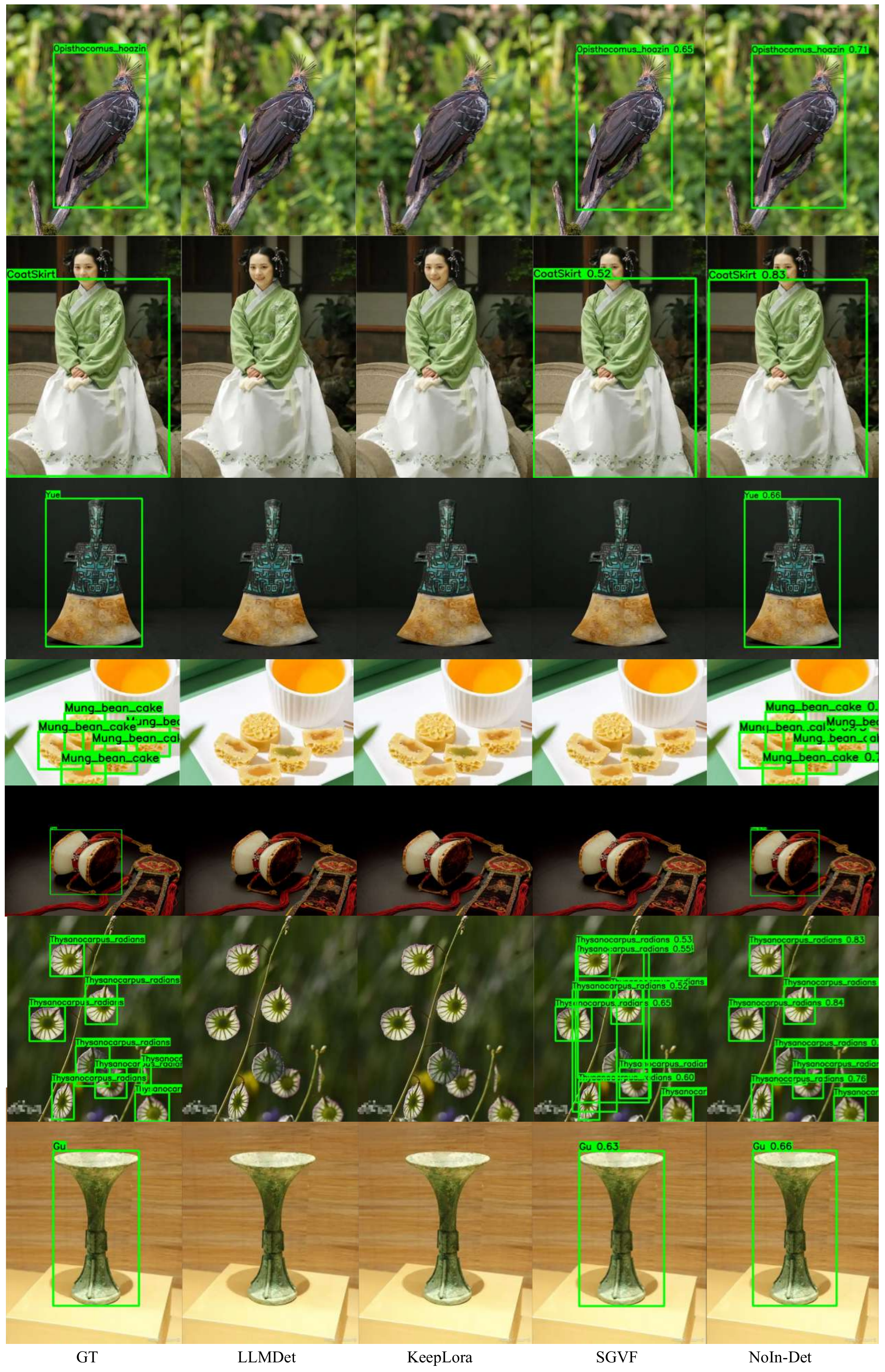}\vspace{-10pt}
	\caption{Qualitative comparison of different methods on Novel-114. Our method produces more accurate and stable detections across various scenarios.}  
	\label{fig:vision}
	\vspace{-10pt}
\end{figure*}

\subsection{Qualitative Results}
\label{sec:Qualitative}
We further provide qualitative comparisons in Fig.~\ref{fig:vision} to validate the effectiveness of our method. Existing methods exhibit different failure patterns under continual novel concept injection. LLMDet and KeepLoRA tend to miss novel objects or produce incomplete detections, indicating insufficient adaptation to newly introduced concepts and unstable retention of historically injected concepts. SGVF shows stronger adaptability to current-task concepts, but often produces redundant detections and false positives, especially on background regions or semantically related objects, suggesting limited stability and knowledge retention.

In contrast, our method produces more accurate and consistent predictions across diverse scenarios. It not only localizes novel objects more precisely and covers object regions more completely, but also effectively avoids spurious or redundant bounding boxes. Its stable behavior across different object scales, complex backgrounds, and visually similar categories further shows that NoIn-Det learns more reliable concept-region alignment rather than relying on unstable task-specific cues. These qualitative results demonstrate that our method better balances plasticity and stability, accurately detecting newly injected concepts while effectively preserving historical knowledge and original open-vocabulary capability.

\end{document}